\documentclass[conference, onecolumn]{IEEEtran}

\IEEEoverridecommandlockouts
\usepackage{amsmath,amssymb,amsfonts}

\usepackage{algorithm, algorithmic}

\usepackage{colortbl}
\usepackage{multirow}
\usepackage{graphicx}
\usepackage{textcomp}
\usepackage{xcolor}
\usepackage{natbib}
\usepackage{hyperref} 
\usepackage{array,booktabs}
\graphicspath{{./}} 
\def\BibTeX{{\rm B\kern-.05em{\sc i\kern-.025em b}\kern-.08em
T\kern-.1667em\lower.7ex\hbox{E}\kern-.125emX}}

\begin{document}

\title{Multi-resolution Interpretation and Diagnostics Tool for Natural Language Classifiers \\
%{\footnotesize How to explain Deep Learning based models' decision in text mining? }
}

\author{\IEEEauthorblockN{Peyman Jalali}
\IEEEauthorblockA{\textit{Corporate Model Risk}\\
\textit{Wells Fargo, US}\\ 
peyman.jalali@wellsfargo.com}
\and
\IEEEauthorblockN{Nengfeng Zhou}
\IEEEauthorblockA{\textit{Corporate Model Risk}\\
\textit{Wells Fargo, US}\\ 
nengfeng.zhou@wellsfargo.com} 
\and
\IEEEauthorblockN{Yufei Yu}
\IEEEauthorblockA{\textit{Corporate Model Risk}\\
\textit{Wells Fargo, US}\\ 
yufei.yu@wellsfargo.com} 

}

\maketitle

\def\mystrut(#1,#2){\vrule height #1pt depth #2pt width 0pt} 
\def\arraystretch{1.3}% 1 is the default, change whatever you need 

\begin{abstract}

Developing explainability methods for Natural Language Processing (NLP) models is a challenging task, for two main reasons. First, the high dimensionality of the data (large number of tokens) results in low coverage and in turn small contributions for the top tokens, compared to the overall model performance. Second, owing to their textual nature, the input variables, after appropriate transformations, are effectively binary (presence or absence of a token in an observation), making the input-output relationship difficult to understand. Common NLP interpretation techniques do not have flexibility in resolution, because they usually operate at word-level and provide fully local (message level) or fully global (over all messages) summaries. The goal of this paper is to create more flexible model explainability summaries by segments of observation or clusters of words that are semantically related to each other. In addition, we introduce a root cause analysis method for NLP models, by analyzing representative False Positive and False Negative examples from different segments. At the end, we illustrate, using a Yelp review data set with three segments (Restaurant, Hotel, and Beauty), that exploiting group/cluster structures in words and/or messages can aid in the interpretation of decisions made by NLP models and can be utilized to assess the model's sensitivity or bias towards gender, syntax, and word meanings.

\end{abstract}


\section{Introduction}

Many Natural Language Classifiers has been developed in recent years. One of the challenges is to understand how these models work and explain their decisions. This will help with identifying potential issues of the models \citep{Lertvittayakumjorn2021survey} and as a result gain more trust in them \citep{ribeiro2016model}. 
In general, the NLP explainability methods can be categorized into two groups: local and global explainability approaches \citep{liu2018model}. \citep{gholizadeh2021model} summarized some of the main methods used to derive local explainability, such as Gradient-Based Sensitivity Analysis \citep{arras2017relevant}, Local Interpretable Model-Agnostic Explanations (LIME) \citep{ribeiro2016model}, Layer-wise Relevance Propagation (LRP) \citep{bach2015pixel}. Most of the NLP explainability methods are post-hoc methods. \citep{madsen2021posthoc} provides a comprehensive survey of the post-hoc interpretability approaches for neural NLP models. 

Local explainability is very helpful in understanding why an NLP model is making a certain prediction, as shown in an example in Table \ref{tab:table1} \citep{gholizadeh2021model}. Since, local explainability only applies to individual observations, it doesn't provide a full picture of the NLP classifier because each observation only contains a very small subset of the tokens used by the NLP classifier. Due to high number of observations, reviewing the local explainabilities of all observations is not practical. One solution may be to randomly select a subset of the observations and review their local explainabilities, which is not effective, because it is likely that the random selection results in missing some important observations. Another solution is to look at some typical examples in each segment, similar to the non-redundant local explanations in \citep{ribeiro2016model}. These representative messages can help us understand the different behaviors in each segment. It was also suggested by \citep{Khanna2019black} and \citep{Han2020subset} to look at local explainability for the False Positive and False Negative predictions. 

{\setlength{\fboxsep}{ 1.5pt} \setlength{\fboxrule}{0pt} \colorbox{white!0} 
%\usepackage{adjustbox}
%\tcbset{width=0.9\textwidth,boxrule=0pt,colback=red,arc=0pt,auto outer arc,left=0pt,right=0pt,boxsep=5pt}

\def\arraystretch{1.9}% 1 is the default, change whatever you need 
%\vspace{0.753in} 
\begin{table*}[h] 
\caption{An example of local explainability. (Red words are those used by the model to identify bad review) \citep{gholizadeh2021model}. }
\centering
\begin{tabular}{|p{14cm}|} 
\hline 
Unfortunately there is \colorbox{red!35.0}{\strut nothing} special about this place My husband got the french dip \colorbox{blue!11.0}{\strut and} myself the mushroom \colorbox{blue!13.0}{\strut panini} Mine was \colorbox{red!24.0}{\strut rather} \colorbox{red!100.0}{\strut disappointing} the mushrooms were minced so tiny and the flavor was semi reminiscent of canned cream of mushroom soup on a \colorbox{blue!14.0}{\strut sandwich} I hate leaving \colorbox{red!32.0}{\strut bad} reviews but it wouldn t help anyone if i lied \colorbox{red!13.0}{\strut sorry} \\

\hline
\end{tabular} 
\label{tab:table1}
\end{table*} 
}

NLP explainability can be done globally at the full data level \citep{liu2018model}. This can be achieved by aggregating the local explainability of all messages which is equivalent to conducting a variable importance analysis of the model. This aggregation is often referred to as global explainability analysis. Global explainability provides us with useful information regarding the words that are contributing the most to the model prediction. However, as we shall discuss later, the high dimensionality of NLP data makes the variable importance less useful since the contribution of each top word in the overall model is very small. In addition, only a very small percentage of messages/observations contain these top words. As such, we might need to review at least the top 100 words to get a good coverage of the observations in the data. The top words should be manually reviewed and it's not always intuitive to decide if a top word is reasonable in the model. Table \ref{tab:varimp1} is an example of variable importance at the overall level. 

To assess the effectiveness of variable importance of an explainability method,\citep{gholizadeh2021model} recommends excluding the top $n$ relevant tokens from the data and using the NLP model to make predictions on an evaluation data, once more. If the model performance dropped significantly, this means these excluded top words are very useful for the NLP model. However, this is usually more useful when we are comparing two different token importance methods. Also, one challenge is that it's hard to judge how much performance drop is an indicator of good model explainability. 

\begin{table*}[htbp] 
\centering
\caption{An example of variable importance at the overall level. (Different methods may give different rankings) \citep{gholizadeh2021model} }
\begin{tabular}{|r|l|l|}
\hline
Rank & Method 1 & Method 2 \\
\hline
1 & mediocre & laid \\
\hline
2 & pathetic & unfortunate \\
\hline
3 & lackluster & lackluster \\
\hline
4 & confusing & disappointing \\
\hline
5 & flavorless & crappy \\
\hline
6 & disappointing & outrageous \\
\hline
7 & poor & pathetic \\
\hline
8 & worse & lacking \\
\hline
9 & shitty & shitty \\
\hline
10 & crappy & mediocre \\
\hline
\end{tabular}%
\label{tab:varimp1}%
\end{table*}

Even with the development of so many NLP explainability methods, it’s still very challenging to fully understand the predictions of the NLP models. The NLP models depend on the words in the text data to predict the probability of some events. Therefore, with tens of thousands of words, each of which as an individual predictor, the dimensionality of the data is very high. As a result, common interpretability methods that are useful in low dimensional machine learning models such as variable importance \citep{Louppe2013vip} \citep{Altmann2010vip}, become less effective for NLP models. The main reason is that the contribution of top words to the overall model performance is not significant due to the high dimensionality of the input space. 
Another challenge for NLP model explainability is the binary nature of input variables (presence or absence of a word in an observation). This makes it difficult to employ another set of popular model interpretability tools, commonly referred to as feature effects methods; for example, Partial Dependent Plots (PDP) \citep{Friedman2001PDP} and Individual Conditional Expectation (ICE) \citep{Goldstein2013ICE}, which can help with understanding the relationship between output and continuous input variables in black box machine learning models. 
It is also difficult to evaluate the effectiveness of NLP model explainability methods which usually requires the involvement of human judgment \citep{Lertvittayakumjorn2021survey} or re-estimation of the model which are both expensive to do. If the model being explained is already a reasonably good model, the NLP explainability may not add much value \citep{Smith2020exp}. In which case, NLP explainability only helps with confirming whether a model is performing reasonably. However, when the model has some significant deficiencies or bugs, NLP explainability can be helpful to find these bugs and improve the model. 

Furthermore, common NLP interpretation techniques do not have flexibility in resolution. They usually operate at word-level and provide fully local (message level) \citep{Hu2018LIMEsup} or fully global (over all messages, similar to the variable importance methods) summaries. In this paper, we propose to analyze NLP models at group or segment levels. Our proposal has two parts:

\begin{enumerate}
\item Model explainability by groups/clusters of messages – Groups based on the natural categories of observations, topic detection using Latent Dirichlet Allocation (LDA), message length, percentage of Out of Vocabulary (OOV)
\item Model explainability by groups/clusters of words: 
\begin{enumerate}
\item Syntax: Part of Speech (POS) tagging
\item Meaning: key dimensions in embedding space (e.g., Gender)
\item Sentiment of words 
\end{enumerate}
\end{enumerate} 

Since in the case of NLP data, the rows are used for the observations (messages) and the columns are used for the individual words, groups/clusters of messages amounts to cluster in the row dimension and groups/clusters of words amounts to clustering in the column dimension.

The reminder of this paper is organized as follows: In section II, we introduce several techniques for grouping/clustering of words and messages in text data; in section III, we illustrate the benefits of these approaches using a Yelp review data set and in section IV we conclude the paper.

\section{Methodology}

\subsection{Model explainability by groups/clusters of messages}

% Interpretation and Summarization by Message Segmentation

%The existing interpretation methods for NLP classifier usually only provide fully local (message level) \citep{Hu2018LIMEsup} or fully global (over all messages) summaries. They do not give a comprehensive view of the NLP classifier due to the unstructured and high dimension nature of the NLP data. 

Analyzing variable importance by groups/clusters of messages can provide additional information regarding Natural Language Classifiers. This can be achieved by aggregating the local explainability of all messages in each segment. Reviewing variable importance across different segments provides us with some benchmarks for comparison purposes. Similarities and differences of variable importance between different segments are all useful information about the NLP model. If the variable importance is different across segments, some further investigation may be done to see if the most important words in each segment are reasonable. 

Considering different ways of clustering/grouping of the data can further help us in evaluating the model fairness. However, assessment of the model fairness usually requires access to some sensitive variables, such as race, gender, age etc., which are either not available or not allowed to be used in modeling process. In some cases, the segment assignment is correlated with sensitive variables where, fairness metrics can be calculated based on the segments. For example, if a complaint model is predicting less number of complaints for an OOV cluster, it could indicate that the model is unfair (not effective in identifying complaint) to non-English speakers. 

In addition, the segmentation information can be used in error root cause analysis, including analyzing FP and FN model outcomes from a testing data. The root cause of FP/FN associated with different segments are likely to be different. From the toy example in Table \ref{tab:rooterror}, we can see that the root cause of false negative error for OOV segment is very different. Without segmentation by OOV segment, this error root cause may not be identified. These patterns in individual segments can help to assess the model weaknesses.

{\setlength{\fboxsep}{ 1.5pt} \setlength{\fboxrule}{0pt} \colorbox{white!0} 
%\usepackage{adjustbox}
%\tcbset{width=0.9\textwidth,boxrule=0pt,colback=red,arc=0pt,auto outer arc,left=0pt,right=0pt,boxsep=5pt}

\begin{table*}[htbp] 
\centering
\caption{ Error root cause analysis by segment: (Bad review is positive class. Red is the positive/bad words identified by model and blue words are negative/good words) (Toy example) }
\begin{tabular}{|p{4.055em}|p{7.055em}|p{18.22em}|p{19.335em}|}
\hline
Segment & Error category & Text & Root Cause \\
\hline
Non OOV & False Positive & “I have never had a \colorbox{red!72.0}{\strut bad} meal or \colorbox{red!72.0}{\strut poor} service at any Ono location I really like there food and service. ” & Model could not handle the negative word “never”. It’s actually a good review. Prediction is bad review. \\
\hline
Non OOV & False Negative & “I ve tried time and time again to like this place but the pizza hmmm is \colorbox{red!72.0}{\strut mediocre} at best the pazookie ice cream cookie is \colorbox{blue!72.0}{\strut great} and so is the atmosphere. ” & Model put too much weight on the word “great” in the sentence. \\
\hline
OOV & False Negative & “No hay nada \colorbox{blue!72.0}{\strut especial} en este lugar. El mío fue bastante decepcionante. ” & Model could not handle OOV word “decepcionante”. The English translation for this is “There is \colorbox{red!72.0}{\strut nothing} special about this place. Mine was rather \colorbox{red!72.0}{\strut disappointing}.” \\
\hline
\end{tabular}%
\label{tab:rooterror}% 
\end{table*} 

}

Many real NLP data sets contain natural segment information. For example, Amazon review data is coming from different categories of products. The product category is a natural segment variable. In absence of a natural segment variable, one can use the following clustering techniques to segment the text data: 

\begin{enumerate}
\item Latent Dirichlet Allocation (LDA) \citep{Blei2003LDA}. LDA is an unsupervised text clustering technique whose purpose is to cluster the messages into a pre-selected number of topics, based on the text information in the messages. Using LDA, one can classify each message to multiple topics or enforce each message to belong to one topic. 

\item Segmentation using summary statistics of the messages, such as message length and \% of Out of Vocabulary (OOV) words. The objective in calculating these statistics is to investigate the text data as well as the model predictions and try to identify potential issues with the actual labels (ground truth) or model predictions. 

\begin{enumerate}
\item {\it message length} might play a role in the labeling process. For example, if the labeling process of a potential complaint message is based on key words match, longer message is more likely to be a complaint since it’s more likely to be matched by the complaint key words. Many Natural Language Classifiers are based on the existence of some signals in the NLP data. The longer the message, the more likely there is some signal. Therefore, message length could also influence the labels predicted by the NLP model. See also \citep{Amplayo2019Length}.

\item {\it OOV} can impact the NLP model performance significantly in the context generation of word embedding \citep{Garneau2019OOT}. Large \% of OOV can also impact the chance of a message being labeled as a complaint. If large \% of OOV is caused by foreign language in a message, this could cause labeling bias for messages from non-English speakers. This bias caused by \% of OOV is likely to impact the model prediction of messages with large \% of OOV. Poor model performance for messages with large \% of OOV can lead to an unfair model for some sub groups of people (non-English speakers) \citep{blodgett2020language}.

\end{enumerate}

\end{enumerate}

\subsection{Model explainability by groups/clusters of words}

Another dimension in the NLP data is the word dimension, which is the columns in the data. Segmentation can also be done in the word dimension. The followings are some examples of segmentation in word dimension: 

\subsubsection{Word segmentation based on meanings of words in NLP data} 

For example, one can segment the data based on some words which are related to sensitive demographic information such as gender or race and check whether the NLP model’s predictions are impacted by such words, in which case model fairness may be challenged.

In order to segment the words based on gender or race, we can use the word embeddings (e.g. word2Vec, glove) which contain information about word meanings such as race and gender. An example of the segmentation based on embeddings is shown in the Table \ref{tab:embedding3}. In this example, embedding similarities of each word are calculated relative to male/female group of words. The male words used to calculate the embedding similarities are ['man' , 'men' , 'he' , 'his', 'sir', 'gentleman'] and the female words used to calculate the embedding similarities are ['woman', 'women', 'she', 'her', 'madam', 'lady']. The Glove embedding (“glove.6B/glove.6B.300d.w2vformat.txt”) of each word is used in the cosine similarity calculation. The average similarity to the male/female words are calculated for each word of interest. Then the difference of the two similarities are calculated. It can be seen that male related words have higher similarities to male than female. Similarly, female related words have higher similarities to female than male. The difference of the male similarity and female similarity can be used to approximate the gender information. This method can be used to automatically find a segmentation of male/female words. 

% Table generated by Excel2LaTeX from sheet 'table3_4'
\begin{table*}[htbp]
\centering
\caption{Meaning segmentation based on embedding similarities to gender words}
\begin{tabular}{|l|r|r|r|c|}
\hline
Token & \multicolumn{1}{c|}{Male similarity} & \multicolumn{1}{c|}{Female similarity} & \multicolumn{1}{c|}{difference} & magnitude \\
\hline
pink & 0.174 & 0.306 & -0.132 & - - - \\
\hline
ballet & 0.147 & 0.259 & -0.112 & - - - \\
\hline
asian & 0.302 & 0.411 & -0.109 & - - - \\
\hline
hispanic & 0.224 & 0.358 & -0.134 & - - - \\
\hline
nurse & 0.237 & 0.45 & -0.213 & - - - - \\
\hline
architect & 0.351 & 0.125 & 0.226 & + + + + \\
\hline
cashier & 0.12 & 0.247 & -0.127 & - - - \\
\hline
player & 0.454 & 0.318 & 0.136 & + + + \\
\hline
singer & 0.266 & 0.381 & -0.115 & - - - \\
\hline
diva & 0.165 & 0.315 & -0.15 & - - - \\
\hline
conductor & 0.268 & 0.118 & 0.15 & + + + \\
\hline
composer & 0.306 & 0.163 & 0.143 & + + + \\
\hline
mob & 0.291 & 0.183 & 0.108 & + + + \\
\hline
thief & 0.429 & 0.306 & 0.123 & + + + \\
\hline
\end{tabular}%
\label{tab:embedding3}%
\end{table*}%

\subsubsection{Word segmentation based on sentiment of words in NLP data}

There are different ways that one can use to get the sentiment of words. It's straightforward to use existing dictionaries for evaluating emotion in text. The tidytext package \citep{Silge2016tidy} provides access to several sentiment lexicons. It includes three general-purpose lexicons: 

\begin{enumerate}
\item AFINN by Finn Årup Nielsen \citep{Nielsen2011afinn}: assigns words with a score (between -5 and 5), with negative scores indicating negative sentiment and positive scores indicating positive sentiment.
\item BING by Bing Liu and collaborators \citep{Bing2012sent}: categorizes words in a binary fashion into positive and negative categories. 

\item NRC by Saif Mohammad and Peter Turney \citep{mohammad2020practical} : categorizes words in a binary fashion (“yes”/“no”) into categories of positive, negative, anger, anticipation, disgust, fear, joy, sadness, surprise, and trust.

\end{enumerate}

All these lexicons are based on single words. In our case study, we choose to use the Bing method since we are only interested in binary emotion and the Bing method has higher match rate than the other two methods in our data. 

\subsubsection{Segmentation of words based on syntax}

Syntax is the arrangement of words and phrases to create well-formed sentences in a language. Part of Speech (POS) tagging is one of the syntax examples. POS tagging is a category of words (or, more generally, of lexical items) that have similar grammatical properties. POS tagging can be implemented automatically through some machine learning model, trained on NLP data with tags. An example of such data set is the universal tagset of NLTK \citep{Bird2009}, which comprises of 12 tag classes: Verb, Noun, Pronouns, Adjectives, Adverbs, Adpositions (prepositions/postpositions), Conjunctions, Determiners, Cardinal Numbers, Particles, Other/ Foreign words, Punctuations. Words that are assigned to the same POS generally display similar syntactic behavior. 
Segmentation based on syntax/POS can help to understand the NLP model better. For example, the local explainability can be combined with the POS segmentation to see which syntax category of the speech has largest impact to the model prediction. The POS segmentation can also be combined with the global explainability to visually show which words are most important in each POS segmentation.

\section{Experiment Results}

The data set used in this paper is provided by Yelp (https://www.kaggle.com/yelp-dataset/yelp-dataset) which includes 720,399 reviews from many topics, including Restaurant, Beauty, Hotel etc. The data was collected between 01/12/2019 and 01/28/2021. The review ratings were from 1 star to 5 stars. In our analysis, we made three exclusions: we excluded the 3 star reviews; we used the data from 12/01/2019 to 01/28/2021; and we limited the topics to three topics: Restaurant, Beauty, and Hotel. After these exclusions, the size of the data dropped to 468,295 reviews. Next, we relabeled the data by assigning label 1 (complaint) to 1 star and 2 stars reviews and label 0 (non-complaint) to 4 stars and 5 stars reviews. To further reduce the size of the data, we randomly selected 15,000 from topic : Restaurant, 10,000 from topic: Beauty and 10,000 from topic: Hotel. In each of the topics the selections were evenly split between 0 and 1 labels, resulting in a completely balanced data set.

To clean the text data, we first split the reviews into lowercase words, filtered out punctuation, removed stop words and used regular expression logic to replace common text-patterns such as number, money, time, phone, date, ssn, url, html, and email with appropriate tags. Then we used the 300d word embedding of these observations (provided by pre-trained vectors of ConceptNet NumberBatch v17.06). We then randomly split the data into 75 percent training and 25 percent testing. 

To classify the data, we used a CNN model with one convolutional layer, maxpooling and ReLU as activation function. For training the model, we completed 10 passes trough the data (epoch = 10) and for each model update, we used 256 samples (batch size = 256). We pre-selected the number of filters at 100, pad size at 100, and used kernel sizes of 1 and 2 with stride size 1. 

Then for model explainability, we used the python package iNNvestigate \citep{alber2019innvestigation} to run Layer-wise Relevance Propagation (LRP)  on both positive records of the training data and positive records of the evaluation data separately. The list of the most influential tokens retrieved by LRP are summarized over the full data and by topics. Note that to filter the rare tokens, we used an adaptive frequency threshold of 0.01\% of total number of observations. A relative threshold based on total number of observations is appropriate for different topics since they have different sample size.

\subsection{Model Performance by Topics}

% Table generated by Excel2LaTeX from sheet 'summary'
\begin{table}[htbp]
\centering
\caption{Model performance by topics (with one model trained on all data).}
\begin{tabular}{|l|cc|cc|cc|cc|}
\hline
& \multicolumn{2}{c|}{All data} & \multicolumn{2}{c|}{Topic: Restaurant} & \multicolumn{2}{c|}{Topic: Hotel} & \multicolumn{2}{c|}{Topic: Beauty} \\
\hline
& \multicolumn{1}{l|}{Pred = 0} & \multicolumn{1}{l|}{Pred = 1} & \multicolumn{1}{l|}{Pred = 0} & \multicolumn{1}{l|}{Pred = 1} & \multicolumn{1}{l|}{Pred = 0} & \multicolumn{1}{l|}{Pred = 1} & \multicolumn{1}{l|}{Pred = 0} & \multicolumn{1}{l|}{Pred = 1} \\
\hline
Actual = 0 & \multicolumn{1}{r|}{3285} & \multicolumn{1}{r|}{195} & \multicolumn{1}{r|}{1425} & \multicolumn{1}{r|}{87} & \multicolumn{1}{r|}{939} & \multicolumn{1}{r|}{45} & \multicolumn{1}{r|}{921} & \multicolumn{1}{r|}{63} \\
\hline
Actual = 1 & \multicolumn{1}{r|}{177} & \multicolumn{1}{r|}{3343} & \multicolumn{1}{r|}{80} & \multicolumn{1}{r|}{1408} & \multicolumn{1}{r|}{41} & \multicolumn{1}{r|}{975} & \multicolumn{1}{r|}{56} & \multicolumn{1}{r|}{960} \\
\hline
F1 Score & \multicolumn{2}{c|}{0.950} & \multicolumn{2}{c|}{0.940} & \multicolumn{2}{c|}{0.960} & \multicolumn{2}{c|}{0.940} \\
\hline
\end{tabular}% 
\label{tab:f1model1}%
\end{table}%

% Table generated by Excel2LaTeX from sheet 'summary'
\begin{table}[htbp]
\centering
\caption{Model performance for each topic (with separate models for each topic). }
\begin{tabular}{|l|ll|ll|ll|ll|}
\hline
& \multicolumn{2}{c|}{All data} & \multicolumn{2}{c|}{Topic: Restaurant} & \multicolumn{2}{c|}{Topic: Hotel} & \multicolumn{2}{c|}{Topic: Beauty} \\
\hline
& \multicolumn{1}{l|}{Pred = 0} & Pred = 1 & \multicolumn{1}{l|}{Pred = 0} & Pred = 1 & \multicolumn{1}{l|}{Pred = 0} & Pred = 1 & \multicolumn{1}{l|}{Pred = 0} & Pred = 1 \\
\hline
Actual = 0 & \multicolumn{1}{l|}{3285} & 195 & \multicolumn{1}{l|}{1451} & 61 & \multicolumn{1}{l|}{912} & 72 & \multicolumn{1}{l|}{922} & 62 \\
\hline
Actual = 1 & \multicolumn{1}{l|}{221} & 3299 & \multicolumn{1}{l|}{123} & 1365 & \multicolumn{1}{l|}{26} & 990 & \multicolumn{1}{l|}{72} & 944 \\
\hline
F1 Score & \multicolumn{2}{l|}{0.950} & \multicolumn{2}{l|}{0.940} & \multicolumn{2}{l|}{0.950} & \multicolumn{2}{l|}{0.930} \\
\hline
\end{tabular}%
\label{tab:f1model4}%
\end{table}%

Table \ref{tab:f1model1} shows the model performance of a model trained on all data from three segments, with performance measured by topics. Table \ref{tab:f1model4} shows the model performance of three separate models, each trained on data from one segment. Fitting a bigger model for all three segments (Table \ref{tab:f1model1}) has slightly better model performance, measured by F1 scores of individual segments. We will further analyze the token importance of individual segments to understand more why a bigger model is preferred rather than fitting three individual models.

% Table generated by Excel2LaTeX from sheet 'summary'
\begin{table}[htbp]
\centering
\caption{ FP/FN rates segmented by the length of messages (Orange cells are the segment with highest error rate for a topic. Blue cells are the segments with lowest error rate.) }
\begin{tabular}{|l|l|l|l|l|l|l|}
\hline
Error Type & Topic & \multicolumn{1}{p{6.28em}|}{len $<$ 20} & \multicolumn{1}{p{6.28em}|}{20 $\leq$ len $<$ 50} & \multicolumn{1}{p{6.28em}|}{50 $\leq$ len $<$ 80} & \multicolumn{1}{p{6.28em}|}{80 $\leq$ len} & \multicolumn{1}{p{6.28em}|}{N of Errors} \\
\hline
FP & Topic: Restaurant & 2.92\% & 2.85\% & \cellcolor[rgb]{ .608, .761, .902} 1.76\% & \cellcolor[rgb]{ 1, .753, 0} 4.48\% & 87 \\
\hline
FP & Topic: Hotel & \cellcolor[rgb]{ 1, .753, 0} 2.70\% & \cellcolor[rgb]{ .608, .761, .902} 1.87\% & 2.63\% & 2.09\% & 44 \\
\hline
FP & Topic: Beauty & \cellcolor[rgb]{ 1, .753, 0} 3.56\% & \cellcolor[rgb]{ .608, .761, .902} 2.59\% & 3.58\% & 3.46\% & 63 \\
\hline
FN & Topic: Restaurant & 2.92\% & 2.47\% & \cellcolor[rgb]{ .608, .761, .902} 2.29\% & \cellcolor[rgb]{ 1, .753, 0} 2.99\% & 78 \\
\hline
FN & Topic: Hotel & 1.20\% & 2.11\% & \cellcolor[rgb]{ .608, .761, .902} 0.96\% & \cellcolor[rgb]{ 1, .753, 0} 3.93\% & 41 \\
\hline
FN & Topic: Beauty & 2.85\% & \cellcolor[rgb]{ .608, .761, .902} 1.94\% & 2.86\% & \cellcolor[rgb]{ 1, .753, 0} 4.04\% & 56 \\
\hline
\end{tabular}%
\label{tab:length}%
\end{table}%

Table \ref{tab:length} shows some analysis of FP/FN rates segmented by the length of messages. It shows some interesting patterns. The model usually performed best when the length is not too long or too short. This makes sense since too short messages may contain too little information to get a good predictions. When a message is too long, it may contain too much un-relevant messages and it could also confuse the models. We will see more evidence in the FP/FN error analysis later. 

\subsection{Token importance by Topics}

LRP methods \citep{bach2015pixel} are used to get the local explainability. Local explainability are aggregated together to get global token importance. This is done at the full data level and topic level.

% Automatic_color_explainability.xlsx _final_pos_neg_sim
\begin{table*}[htbp]
\centering
\caption{Token importance contributing to complaints, summarized by topics (ranked by embedding similarity to top words). The embedding similarity ranks are reflected with colors (Blue words have higher similarity ranks). }
\begin{tabular}{|r|l|l|l|l|}
\hline
\rowcolor[rgb]{ .675, .725, .792} \multicolumn{1}{|p{4.055em}|}{\textbf{Token Rank}} & \multicolumn{1}{p{8.22em}|}{\textbf{full\_data}} & \multicolumn{1}{p{7.78em}|}{\textbf{topic: Restaurant}} & \multicolumn{1}{p{7.72em}|}{\textbf{topic: Hotel}} & \multicolumn{1}{p{9.165em}|}{\textbf{topic: Beauty}} \\
\hline
0 & \cellcolor[rgb]{ .608, .761, .902} misleading & \cellcolor[rgb]{ .608, .761, .902} incomplete & \cellcolor[rgb]{ .608, .761, .902} disappointing & \cellcolor[rgb]{ .608, .761, .902} worst \\
\hline
1 & \cellcolor[rgb]{ .608, .761, .902} tasteless & \cellcolor[rgb]{ .608, .761, .902} miserable & \cellcolor[rgb]{ .608, .761, .902} worst & \cellcolor[rgb]{ .608, .761, .902} disappointing \\
\hline
2 & \cellcolor[rgb]{ .608, .761, .902} miserable & \cellcolor[rgb]{ .608, .761, .902} tasteless & \cellcolor[rgb]{ .608, .761, .902} horrible & \cellcolor[rgb]{ .608, .761, .902} rude \\
\hline
3 & \cellcolor[rgb]{ .608, .761, .902} worst & \cellcolor[rgb]{ .608, .761, .902} worst & \cellcolor[rgb]{ .741, .843, .933} unacceptable & \cellcolor[rgb]{ .608, .761, .902} horrible \\
\hline
4 & \cellcolor[rgb]{ .608, .761, .902} disappointing & \cellcolor[rgb]{ .608, .761, .902} embarrassing & \cellcolor[rgb]{ .608, .761, .902} rude & \cellcolor[rgb]{ .608, .761, .902} disgusting \\
\hline
5 & \cellcolor[rgb]{ .608, .761, .902} embarrassing & \cellcolor[rgb]{ .608, .761, .902} disappointing & \cellcolor[rgb]{ .867, .922, .969} poor & \cellcolor[rgb]{ .608, .761, .902} dishonest \\
\hline
6 & \cellcolor[rgb]{ .608, .761, .902} rude & \cellcolor[rgb]{ .608, .761, .902} horrible & \cellcolor[rgb]{ .608, .761, .902} worse & \cellcolor[rgb]{ .608, .761, .902} filthy \\
\hline
7 & \cellcolor[rgb]{ .608, .761, .902} horrible & \cellcolor[rgb]{ .608, .761, .902} disgusting & \cellcolor[rgb]{ .741, .843, .933} unprofessional & \cellcolor[rgb]{ .867, .922, .969} poor \\
\hline
8 & \cellcolor[rgb]{ .608, .761, .902} horrendous & \cellcolor[rgb]{ .608, .761, .902} rude & \cellcolor[rgb]{ .608, .761, .902} terrible & \cellcolor[rgb]{ .608, .761, .902} outdated \\
\hline
9 & \cellcolor[rgb]{ .608, .761, .902} horribly & \cellcolor[rgb]{ .608, .761, .902} horrendous & \cellcolor[rgb]{ .867, .922, .969} sloppy & \cellcolor[rgb]{ .741, .843, .933} unacceptable \\
\hline
10 & \cellcolor[rgb]{ .741, .843, .933} incompetent & \cellcolor[rgb]{ .867, .922, .969} rotten & \cellcolor[rgb]{ .608, .761, .902} disappointed & \cellcolor[rgb]{ .608, .761, .902} worse \\
\hline
11 & \cellcolor[rgb]{ .741, .843, .933} unsanitary & \cellcolor[rgb]{ .741, .843, .933} incompetent & \cellcolor[rgb]{ .741, .843, .933} waste & \cellcolor[rgb]{ .608, .761, .902} terrible \\
\hline
12 & \cellcolor[rgb]{ .608, .761, .902} disgusting & \cellcolor[rgb]{ .608, .761, .902} pathetic & \cellcolor[rgb]{ .867, .922, .969} sad & \cellcolor[rgb]{ .741, .843, .933} unprofessional \\
\hline
13 & \cellcolor[rgb]{ .867, .922, .969} poor & \cellcolor[rgb]{ .608, .761, .902} horribly & \cellcolor[rgb]{ .741, .843, .933} dirty & \cellcolor[rgb]{ .741, .843, .933} waste \\
\hline
14 & \cellcolor[rgb]{ .608, .761, .902} dishonest & \cellcolor[rgb]{ .867, .922, .969} poor & \cellcolor[rgb]{ .608, .761, .902} disrespectful & \cellcolor[rgb]{ .608, .761, .902} disappointed \\
\hline
15 & \cellcolor[rgb]{ .608, .761, .902} horrific & \cellcolor[rgb]{ .608, .761, .902} racist & \cellcolor[rgb]{ .867, .922, .969} no & \cellcolor[rgb]{ .867, .922, .969} ridiculous \\
\hline
16 & \cellcolor[rgb]{ .741, .843, .933} unacceptable & \cellcolor[rgb]{ .741, .843, .933} unacceptable & \cellcolor[rgb]{ .988, .894, .839} smell & \cellcolor[rgb]{ .867, .922, .969} no \\
\hline
17 & \cellcolor[rgb]{ .988, .894, .839} undercooked & \cellcolor[rgb]{ .988, .894, .839} undercooked & \cellcolor[rgb]{ .867, .922, .969} dollars & \cellcolor[rgb]{ .741, .843, .933} unhelpful \\
\hline
18 & \cellcolor[rgb]{ .608, .761, .902} pathetic & \cellcolor[rgb]{ .608, .761, .902} filthy & \cellcolor[rgb]{ .867, .922, .969} not & \cellcolor[rgb]{ .608, .761, .902} lied \\
\hline
19 & \cellcolor[rgb]{ .741, .843, .933} soggy & \cellcolor[rgb]{ .741, .843, .933} soggy & \cellcolor[rgb]{ .867, .922, .969} ridiculous & \cellcolor[rgb]{ .741, .843, .933} dirty \\
\hline
20 & \cellcolor[rgb]{ .608, .761, .902} racist & \cellcolor[rgb]{ .608, .761, .902} terrible & \cellcolor[rgb]{ .988, .894, .839} management & \cellcolor[rgb]{ .608, .761, .902} charges \\
\hline
\end{tabular}%
\label{tab:comp1}%
\end{table*}%

\subsubsection{Token importance ranked by meaning of words} 
Table \ref{tab:comp1} shows the token importance by topics (contributing to complaints). Due to the limit of the space, only the top 20 words are shown in the table. We also reviewed the top 100 words in this analysis. The important tokens are ranked by embedding similarity difference, similarly to the methods as shown in Table \ref{tab:embedding3}. Instead of using gender words, the embedding similarities are calculated as similarity to top 10 complaints words and top 10 non-complaints words of the full data model. 

We can see that most of the top 20 words have same colors, which means they fall in the same category based on embedding similarity ranks. This is true across the different topic categories. There are only a few exceptions such as “undercooked”, “smell”, and “management”. These words are related to their respective category, “undercooked” for Restaurants, “Smell” for Beauty, and “management” for Beauty and Hotels. If we compare the top words across different categories, we can see different categories share many top complaint words. This partly explains why fitting a bigger model with all three segments have the best model performance. More details on the unique top words in a topic can be found in Table \ref{tab:uniq}.

Out of the top 100 tokens, there are some additional exceptions. All the exceptions from words ranking 21 to 60 are ``minutes'', ``overpriced'', ``st'', ``told'', ``business'', ``chipping'', ``oh'', ``paid'', ``dry'', ``hour''. Most of these tokens are reasonable, and related to their respective category. The percentage of exceptions from words with rankings between 61 and 100 is slightly higher. However, they have less impact on the model prediction due to their higher token ranking.

% Automatic_color_explainability.xlsx _final_pos_neg_sim
\begin{table}[htbp]
\centering
\caption{ Token importance contributing to non-complaints, summarized by topics (ranked by embedding similarity to top words). The embedding similarity ranks are reflected with colors (Orange words have higher similarity ranks). }
\begin{tabular}{|r|l|l|l|l|}
\hline
\rowcolor[rgb]{ .675, .725, .792} \multicolumn{1}{|p{4.055em}|}{\textbf{Token Rank}} & \multicolumn{1}{p{8.165em}|}{\textbf{full\_data}} & \multicolumn{1}{p{8.165em}|}{\textbf{topic: Restaurant}} & \multicolumn{1}{p{8.165em}|}{\textbf{topic: Hotel}} & \multicolumn{1}{p{8.165em}|}{\textbf{topic: Beauty}} \\
\hline
0 & \cellcolor[rgb]{ .957, .69, .518} delicious & \cellcolor[rgb]{ .957, .69, .518} delicious & \cellcolor[rgb]{ .957, .69, .518} amazing & \cellcolor[rgb]{ .957, .69, .518} delicious \\
\hline
1 & \cellcolor[rgb]{ .957, .69, .518} amazing & \cellcolor[rgb]{ .957, .69, .518} amazing & \cellcolor[rgb]{ .957, .69, .518} wonderful & \cellcolor[rgb]{ .957, .69, .518} amazing \\
\hline
2 & \cellcolor[rgb]{ .957, .69, .518} superb & \cellcolor[rgb]{ .957, .69, .518} superb & \cellcolor[rgb]{ .957, .69, .518} beautiful & \cellcolor[rgb]{ .957, .69, .518} wonderful \\
\hline
3 & \cellcolor[rgb]{ .957, .69, .518} thoughtful & \cellcolor[rgb]{ .957, .69, .518} thoughtful & \cellcolor[rgb]{ .957, .69, .518} incredible & \cellcolor[rgb]{ .957, .69, .518} informative \\
\hline
4 & \cellcolor[rgb]{ .957, .69, .518} gracious & \cellcolor[rgb]{ .957, .69, .518} tasty & \cellcolor[rgb]{ .957, .69, .518} excellent & \cellcolor[rgb]{ .957, .69, .518} excellent \\
\hline
5 & \cellcolor[rgb]{ .957, .69, .518} wonderful & \cellcolor[rgb]{ .957, .69, .518} gracious & \cellcolor[rgb]{ .957, .69, .518} fabulous & \cellcolor[rgb]{ .957, .69, .518} fantastic \\
\hline
6 & \cellcolor[rgb]{ .957, .69, .518} tasty & \cellcolor[rgb]{ .957, .69, .518} incredible & \cellcolor[rgb]{ .973, .796, .678} best & \cellcolor[rgb]{ .957, .69, .518} incredible \\
\hline
7 & \cellcolor[rgb]{ .957, .69, .518} incredible & \cellcolor[rgb]{ .957, .69, .518} wonderful & \cellcolor[rgb]{ .957, .69, .518} fantastic & \cellcolor[rgb]{ .957, .69, .518} beautiful \\
\hline
8 & \cellcolor[rgb]{ .957, .69, .518} beautiful & \cellcolor[rgb]{ .957, .69, .518} def & \cellcolor[rgb]{ .957, .69, .518} perfect & \cellcolor[rgb]{ .973, .796, .678} best \\
\hline
9 & \cellcolor[rgb]{ .957, .69, .518} informative & \cellcolor[rgb]{ .957, .69, .518} informative & \cellcolor[rgb]{ .957, .69, .518} great & \cellcolor[rgb]{ .957, .69, .518} perfect \\
\hline
10 & \cellcolor[rgb]{ .957, .69, .518} def & \cellcolor[rgb]{ .957, .69, .518} excellent & \cellcolor[rgb]{ .957, .69, .518} awesome & \cellcolor[rgb]{ .957, .69, .518} enjoyable \\
\hline
11 & \cellcolor[rgb]{ .957, .69, .518} excellent & \cellcolor[rgb]{ .957, .69, .518} fantastic & \cellcolor[rgb]{ .957, .69, .518} meticulous & \cellcolor[rgb]{ .957, .69, .518} fabulous \\
\hline
12 & \cellcolor[rgb]{ .957, .69, .518} fantastic & \cellcolor[rgb]{ .957, .69, .518} fabulous & \cellcolor[rgb]{ .957, .69, .518} talented & \cellcolor[rgb]{ .957, .69, .518} great \\
\hline
13 & \cellcolor[rgb]{ .957, .69, .518} fabulous & \cellcolor[rgb]{ .973, .796, .678} best & \cellcolor[rgb]{ .957, .69, .518} enjoyed & \cellcolor[rgb]{ .973, .796, .678} affordable \\
\hline
14 & \cellcolor[rgb]{ .973, .796, .678} best & \cellcolor[rgb]{ .957, .69, .518} beautiful & \cellcolor[rgb]{ .988, .894, .839} clean & \cellcolor[rgb]{ .957, .69, .518} awesome \\
\hline
15 & \cellcolor[rgb]{ .957, .69, .518} flavorful & \cellcolor[rgb]{ .957, .69, .518} yummy & \cellcolor[rgb]{ .973, .796, .678} appreciated & \cellcolor[rgb]{ .973, .796, .678} professional \\
\hline
16 & \cellcolor[rgb]{ .957, .69, .518} yummy & \cellcolor[rgb]{ .957, .69, .518} flavorful & \cellcolor[rgb]{ .973, .796, .678} professional & \cellcolor[rgb]{ .973, .796, .678} easy \\
\hline
17 & \cellcolor[rgb]{ .973, .796, .678} timely & \cellcolor[rgb]{ .957, .69, .518} great & \cellcolor[rgb]{ .957, .69, .518} personable & \cellcolor[rgb]{ .957, .69, .518} lovely \\
\hline
18 & \cellcolor[rgb]{ .957, .69, .518} wonderfully & \cellcolor[rgb]{ .957, .69, .518} awesome & \cellcolor[rgb]{ .957, .69, .518} lovely & \cellcolor[rgb]{ .988, .894, .839} clean \\
\hline
19 & \cellcolor[rgb]{ .957, .69, .518} refreshing & \cellcolor[rgb]{ .957, .69, .518} enjoyable & \cellcolor[rgb]{ .973, .796, .678} easy & \cellcolor[rgb]{ .957, .69, .518} spacious \\
\hline
20 & \cellcolor[rgb]{ .957, .69, .518} entertaining & \cellcolor[rgb]{ .957, .69, .518} perfect & \cellcolor[rgb]{ .957, .69, .518} grateful & \cellcolor[rgb]{ .957, .69, .518} knowledgeable \\
\hline
\end{tabular}%
\label{tab:noncomp1}%
\end{table}%

Table \ref{tab:noncomp1} shows the token importance by topics (contributing to non-complaints), ranked by the embedding similarity difference method. Overall, the non-complaints words are more similar across different categories. More details on the unique top words in a topic can be found in Table \ref{tab:uniq}. They also have more consistently embedding similarity ranks (reflected by colors). We can see that all of the top 20 words have same colors, which means they fall in the same category based on embedding similarities. This is true across the different topic categories. 
Out of top 60 words (due to space limitation, only 20 is shown here), we only observed one exception word ``overall'' with different color. The percentage of exceptions from words ranking 61 to 100 is slightly higher. However, they have less impact to the model prediction due to their higher ranking number. Interestingly, the exceptions words in Hotel category are all related to city names or location, such as ``orlando'', ``town'', ``city'', ``atlanta'', ``boston''.

\subsubsection{Token importance ranked by sentiment of words}

% Table generated by Excel2LaTeX from sheet '_final_pos_neg_sim' Automatic_color_explainability - Bing.xlsx
\begin{table}[htbp]
\centering
\caption{Token importance contributing to complaints, summarized by topics (ranked by Bing sentiment, See https://www.tidytextmining.com/sentiment.html 
). The sentiment ranks are reflected with colors (Blue words have negative sentiment ranks). Rows with all blues are removed to save space (refer to Table \ref{tab:comp1} for those tokens). }
\begin{tabular}{|r|l|l|l|l|}
\hline
\rowcolor[rgb]{ .675, .725, .792} \multicolumn{1}{|p{4.055em}|}{\textbf{Token Rank}} & \multicolumn{1}{p{8.22em}|}{\textbf{full\_data}} & \multicolumn{1}{p{7.78em}|}{\textbf{topic: Restaurant}} & \multicolumn{1}{p{7.72em}|}{\textbf{topic: Hotel}} & \multicolumn{1}{p{9.165em}|}{\textbf{topic: Beauty}} \\
\hline
0 & \cellcolor[rgb]{ .608, .761, .902} misleading & \cellcolor[rgb]{ .608, .761, .902} incomplete & \cellcolor[rgb]{ .608, .761, .902} disappointing & \cellcolor[rgb]{ .608, .761, .902} worst \\
\hline
1 & tasteless & \cellcolor[rgb]{ .608, .761, .902} miserable & \cellcolor[rgb]{ .608, .761, .902} worst & \cellcolor[rgb]{ .608, .761, .902} disappointing \\
\hline
2 & \cellcolor[rgb]{ .608, .761, .902} miserable & tasteless & \cellcolor[rgb]{ .608, .761, .902} horrible & \cellcolor[rgb]{ .608, .761, .902} rude \\
\hline 
7 & \cellcolor[rgb]{ .608, .761, .902} horrible & \cellcolor[rgb]{ .608, .761, .902} disgusting & unprofessional & \cellcolor[rgb]{ .608, .761, .902} poor \\
\hline
8 & \cellcolor[rgb]{ .608, .761, .902} horrendous & \cellcolor[rgb]{ .608, .761, .902} rude & \cellcolor[rgb]{ .608, .761, .902} terrible & outdated \\
\hline
9 & horribly & \cellcolor[rgb]{ .608, .761, .902} horrendous & \cellcolor[rgb]{ .608, .761, .902} sloppy & \cellcolor[rgb]{ .608, .761, .902} unacceptable \\
\hline 
11 & unsanitary & \cellcolor[rgb]{ .608, .761, .902} incompetent & \cellcolor[rgb]{ .608, .761, .902} waste & \cellcolor[rgb]{ .608, .761, .902} terrible \\
\hline
12 & \cellcolor[rgb]{ .608, .761, .902} disgusting & \cellcolor[rgb]{ .608, .761, .902} pathetic & \cellcolor[rgb]{ .608, .761, .902} sad & unprofessional \\
\hline
13 & \cellcolor[rgb]{ .608, .761, .902} poor & horribly & \cellcolor[rgb]{ .608, .761, .902} dirty & \cellcolor[rgb]{ .608, .761, .902} waste \\
\hline 
15 & \cellcolor[rgb]{ .608, .761, .902} horrific & \cellcolor[rgb]{ .608, .761, .902} racist & no & \cellcolor[rgb]{ .608, .761, .902} ridiculous \\
\hline
16 & \cellcolor[rgb]{ .608, .761, .902} unacceptable & \cellcolor[rgb]{ .608, .761, .902} unacceptable & \cellcolor[rgb]{ .608, .761, .902} smell & no \\
\hline
17 & undercooked & undercooked & dollars & \cellcolor[rgb]{ .608, .761, .902} unhelpful \\
\hline
18 & \cellcolor[rgb]{ .608, .761, .902} pathetic & \cellcolor[rgb]{ .608, .761, .902} filthy & not & \cellcolor[rgb]{ .608, .761, .902} lied \\
\hline
19 & soggy & soggy & \cellcolor[rgb]{ .608, .761, .902} ridiculous & \cellcolor[rgb]{ .608, .761, .902} dirty \\
\hline
20 & \cellcolor[rgb]{ .608, .761, .902} racist & \cellcolor[rgb]{ .608, .761, .902} terrible & management & charges \\
\hline
\end{tabular}%
\label{tab:comp2}%
\end{table}%

Table \ref{tab:comp2} shows the token importance by topics (contributing to complaints) with different ranking method. It’s ranked by Bing sentiment (https://www.tidytextmining.com/sentiment.html). Some of the words cannot be matched by the Bing method and they are not colored. We can see that all the top 20 words matched by Bing method have the same color (blue color reflecting negative sentiment ranks). This means they have the same sentiments based on Bing method. Out of the top 100 words, the only word with different color is word `refund', which ranked number 37 in Hotel topic and ranked 58 in Beauty topic. 

For the words which cannot be matched by the Bing method, they are usually topic related words, such as ``tasteless', ``undercooked'', ``soggy'', ``inedible'', ``oily'' in the Restaurant topic. Some other top ranking non-match words are ``unprofessional'', ``outdated'', ``horribly'', ``no'', ``dollars'', ``not'', ``charges'' and ``management''. All these words are reasonable complaint words. The percentage of words not matched by Bing increased slightly with the increase of ranking number from 20 to 100. These non-match words are less intuitive complaint words. However, their impact are less due to their higher ranking number.

Table \ref{tab:noncomp2} shows the token importance by topics (contributing to non-complaints) ranked by Bing sentiment method. We can see that all the top 20 tokens which are matched by Bing method have the same color (orange color reflecting positive sentiment ranks). 
Out of the top 100 tokens (due to space limitation, only 20 is shown here), the first token with different color is `unbelievable', which ranked number 26 in full data and ranked 34 in Restaurant topic. The only other two tokens with different colors are `bomb' and `issues', which have token rankings higher than 60. 
For the words which cannot be matched by the Bing method, they are usually topic related words, such as ``tasty', ``yummy'', ``flavorful'' in the Restaurant topic. Some other top ranking non-match words are ``def'', ``informative'', ``professional'', ``personable''. Most of these words are reasonable non-complaint words. Similarly as the complaint words, the percentage of words not matched by Bing increased slightly with the increase of ranking number. These non-match words are less intuitive non-complaint words. However, their impact are less due to their higher ranking number.

% Table generated by Excel2LaTeX from sheet '_final_pos_neg_sim' Automatic_color_explainability - Bing.xlsx
\begin{table}[htbp]
\centering
\caption{Token importance contributing to non-complaints, summarized by topics (ranked by Bing sentiment, See https://www.tidytextmining.com/sentiment.html 
). The sentiment ranks are reflected with colors (Oranges words have positive sentiment ranks). Rows with all oranges are removed to save space (refer to Table \ref{tab:noncomp1} for those tokens).}
\begin{tabular}{|r|l|l|l|l|} 
\hline
\rowcolor[rgb]{ .675, .725, .792} \multicolumn{1}{|p{4.11em}|}{\textbf{Token Rank}} & \multicolumn{1}{p{7.945em}|}{\textbf{full\_data}} & \multicolumn{1}{p{7.945em}|}{\textbf{topic: Restaurant}} & \multicolumn{1}{p{7.945em}|}{\textbf{topic: Hotel}} & \multicolumn{1}{p{7.945em}|}{\textbf{topic: Beauty}} \\
\hline 
3 & \cellcolor[rgb]{ .957, .69, .518} thoughtful & \cellcolor[rgb]{ .957, .69, .518} thoughtful & \cellcolor[rgb]{ .957, .69, .518} incredible & informative \\
\hline
4 & \cellcolor[rgb]{ .957, .69, .518} gracious & tasty & \cellcolor[rgb]{ .957, .69, .518} excellent & \cellcolor[rgb]{ .957, .69, .518} excellent \\
\hline 
6 & tasty & \cellcolor[rgb]{ .957, .69, .518} incredible & \cellcolor[rgb]{ .957, .69, .518} best & \cellcolor[rgb]{ .957, .69, .518} incredible \\
\hline
7 & \cellcolor[rgb]{ .957, .69, .518} incredible & \cellcolor[rgb]{ .957, .69, .518} wonderful & \cellcolor[rgb]{ .957, .69, .518} fantastic & \cellcolor[rgb]{ .957, .69, .518} beautiful \\
\hline
8 & \cellcolor[rgb]{ .957, .69, .518} beautiful & def & \cellcolor[rgb]{ .957, .69, .518} perfect & \cellcolor[rgb]{ .957, .69, .518} best \\
\hline
9 & informative & informative & \cellcolor[rgb]{ .957, .69, .518} great & \cellcolor[rgb]{ .957, .69, .518} perfect \\
\hline
10 & def & \cellcolor[rgb]{ .957, .69, .518} excellent & \cellcolor[rgb]{ .957, .69, .518} awesome & \cellcolor[rgb]{ .957, .69, .518} enjoyable \\
\hline 
15 & flavorful & yummy & \cellcolor[rgb]{ .957, .69, .518} appreciated & professional \\
\hline
16 & yummy & flavorful & professional & \cellcolor[rgb]{ .957, .69, .518} easy \\
\hline
17 & \cellcolor[rgb]{ .957, .69, .518} timely & \cellcolor[rgb]{ .957, .69, .518} great & personable & \cellcolor[rgb]{ .957, .69, .518} lovely \\
\hline 
20 & \cellcolor[rgb]{ .957, .69, .518} entertaining & \cellcolor[rgb]{ .957, .69, .518} perfect & \cellcolor[rgb]{ .957, .69, .518} grateful & \cellcolor[rgb]{ .957, .69, .518} knowledgeable \\
\hline
\end{tabular}% 
\label{tab:noncomp2}%
\end{table}%

\subsubsection{Token importance ranked by Part of Speech (POS) of words} 

Table \ref{tab:tablepos} shows a summary of part of speech for 800 tokens, from the top 100 tokens (both complaint and non-complaint) of all data and three topics. Here the POS of top token is obtained using a mapping table (https://www.classace.io/tools/part-of-speech-identifier). The results obtained from this method is very reliable with a few exceptions. The colors from Table \ref{tab:tablepos} will be applied to the explainability tables. Some POS with very small number of tokens are grouped together with a larger similar POS. 

We can see that about half of the top tokens are coming from adjective category and this makes sense to us. Adverb and Past participle also has significant number of top tokens. They are relatively close to Adjective. Noun appeared to be the most different category and orange color is used instead of blue color to differentiate.

% Automatic_color_explainability.xlsx pos_pivot
\begin{table}[htbp]
\centering
\caption{Summary of part of speech for 800 tokens, from the top 100 tokens (both complaint and non-complaint) of all data and three topics. The coloring in the table will be applied to Table \ref{tab:pos1} and Table \ref{tab:pos2}. }
\begin{tabular}{|l|r|}
\hline
POS & \multicolumn{1}{l|}{count of token} \\
\hline
\rowcolor[rgb]{ .608, .761, .902} Adjective & \cellcolor[rgb]{ 1, 1, 1} 397 \\
\hline
\rowcolor[rgb]{ .608, .761, .902} Comparative adjective & \cellcolor[rgb]{ 1, 1, 1} 6 \\
\hline
\rowcolor[rgb]{ .608, .761, .902} Superlative adjective & \cellcolor[rgb]{ 1, 1, 1} 11 \\
\hline
\rowcolor[rgb]{ .741, .843, .933} Adverb & \cellcolor[rgb]{ 1, 1, 1} 63 \\
\hline
\rowcolor[rgb]{ .867, .922, .969} Determiner & \cellcolor[rgb]{ 1, 1, 1} 4 \\
\hline
\rowcolor[rgb]{ .867, .922, .969} Interjection & \cellcolor[rgb]{ 1, 1, 1} 2 \\
\hline
\rowcolor[rgb]{ .867, .922, .969} Past participle & \cellcolor[rgb]{ 1, 1, 1} 51 \\
\hline
\rowcolor[rgb]{ .867, .922, .969} Past tense verb & \cellcolor[rgb]{ 1, 1, 1} 34 \\
\hline
\rowcolor[rgb]{ .867, .922, .969} Present tense verb & \cellcolor[rgb]{ 1, 1, 1} 2 \\
\hline
\rowcolor[rgb]{ .867, .922, .969} Verb & \cellcolor[rgb]{ 1, 1, 1} 27 \\
\hline
\rowcolor[rgb]{ .867, .922, .969} Verb gerund & \cellcolor[rgb]{ 1, 1, 1} 25 \\
\hline
\rowcolor[rgb]{ .973, .796, .678} Plural noun & \cellcolor[rgb]{ 1, 1, 1} 49 \\
\hline
\rowcolor[rgb]{ .957, .69, .518} Noun & \cellcolor[rgb]{ 1, 1, 1} 129 \\
\hline
\end{tabular}%
\label{tab:tablepos}%
\end{table}%

Table \ref{tab:pos1} shows the top words contributing to complaints (coloring with POS). The rows with all dark blue (Adjective) are removed to save space. We can see rows ranking from 0 to 8 are all adjective. Adjective words are dominant in the top 30 tokens. The percentage of adjective decreases when the ranking number increases from 30 to 100. Overall, the POS of the words tells us that the model is mainly depending on adjective for predictions.

The percentage of Nouns in top 30 tokens is pretty small. Some words are mistakenly identified as noun even they are adjective, including ``unsanitary'', ``disrespectful''. Some words can be both adjective and noun, such as ``waste''. The other nouns in top 30 words are mostly reasonable complaint related words, such as ``smell'', ``dollars'', ``management'', ``minutes'', ``money'', ``mess''. The percentage of Nouns increase when the ranking number increases from 30 to 100. Generally, the Noun words are less intuitive than adjective with same ranking. Nouns are also a significant part of the model even they can be less intuitive sometimes, especially when the ranking number is high. 

% Automatic_color_explainability.xlsx _final_pos_neg_pos
\begin{table}[htbp]
\centering
\caption{Token importance by topics (colored by POS with colors defined in Table \ref{tab:tablepos} ). Words contributing to complaints. Rows with all dark blue (Adjective) are removed to save space (refer to Table \ref{tab:comp1} for those tokens).}
\begin{tabular}{|r|l|l|l|l|}
\hline
\rowcolor[rgb]{ .675, .725, .792} \multicolumn{1}{|p{4.055em}|}{\textbf{Rank}} & \multicolumn{1}{p{8.22em}|}{\textbf{full\_data}} & \multicolumn{1}{p{7.78em}|}{\textbf{topic: Restaurant}} & \multicolumn{1}{p{7.72em}|}{\textbf{topic: Hotel}} & \multicolumn{1}{p{9.165em}|}{\textbf{topic: Beauty}} \\
\hline
9 & \cellcolor[rgb]{ .741, .843, .933} horribly & \cellcolor[rgb]{ .608, .761, .902} horrendous & \cellcolor[rgb]{ .608, .761, .902} sloppy & \cellcolor[rgb]{ .608, .761, .902} unacceptable \\
\hline
10 & \cellcolor[rgb]{ .608, .761, .902} incompetent & \cellcolor[rgb]{ .608, .761, .902} rotten & \cellcolor[rgb]{ .867, .922, .969} disappointed & \cellcolor[rgb]{ .608, .761, .902} worse \\
\hline
11 & \cellcolor[rgb]{ .957, .69, .518} unsanitary & \cellcolor[rgb]{ .608, .761, .902} incompetent & \cellcolor[rgb]{ .957, .69, .518} waste & \cellcolor[rgb]{ .608, .761, .902} terrible \\
\hline 
13 & \cellcolor[rgb]{ .608, .761, .902} poor & \cellcolor[rgb]{ .741, .843, .933} horribly & \cellcolor[rgb]{ .608, .761, .902} dirty & \cellcolor[rgb]{ .957, .69, .518} waste \\
\hline
14 & \cellcolor[rgb]{ .608, .761, .902} dishonest & \cellcolor[rgb]{ .608, .761, .902} poor & \cellcolor[rgb]{ .957, .69, .518} disrespectful & \cellcolor[rgb]{ .867, .922, .969} disappointed \\
\hline
15 & \cellcolor[rgb]{ .608, .761, .902} horrific & \cellcolor[rgb]{ .608, .761, .902} racist & \cellcolor[rgb]{ .867, .922, .969} no & \cellcolor[rgb]{ .608, .761, .902} ridiculous \\
\hline
16 & \cellcolor[rgb]{ .608, .761, .902} unacceptable & \cellcolor[rgb]{ .608, .761, .902} unacceptable & \cellcolor[rgb]{ .957, .69, .518} smell & \cellcolor[rgb]{ .867, .922, .969} no \\
\hline
17 & \cellcolor[rgb]{ .867, .922, .969} undercooked & \cellcolor[rgb]{ .867, .922, .969} undercooked & \cellcolor[rgb]{ .973, .796, .678} dollars & \cellcolor[rgb]{ .608, .761, .902} unhelpful \\
\hline
18 & \cellcolor[rgb]{ .608, .761, .902} pathetic & \cellcolor[rgb]{ .608, .761, .902} filthy & \cellcolor[rgb]{ .741, .843, .933} not & \cellcolor[rgb]{ .867, .922, .969} lied \\
\hline 
20 & \cellcolor[rgb]{ .608, .761, .902} racist & \cellcolor[rgb]{ .608, .761, .902} terrible & \cellcolor[rgb]{ .957, .69, .518} management & \cellcolor[rgb]{ .973, .796, .678} charges \\
\hline
21 & \cellcolor[rgb]{ .608, .761, .902} filthy & \cellcolor[rgb]{ .608, .761, .902} worse & \cellcolor[rgb]{ .973, .796, .678} minutes & \cellcolor[rgb]{ .957, .69, .518} management \\
\hline
22 & \cellcolor[rgb]{ .608, .761, .902} worse & \cellcolor[rgb]{ .608, .761, .902} runny & \cellcolor[rgb]{ .957, .69, .518} money & \cellcolor[rgb]{ .867, .922, .969} overpriced \\
\hline
23 & \cellcolor[rgb]{ .608, .761, .902} terrible & \cellcolor[rgb]{ .608, .761, .902} lackluster & \cellcolor[rgb]{ .867, .922, .969} said & \cellcolor[rgb]{ .608, .761, .902} gross \\
\hline
24 & \cellcolor[rgb]{ .608, .761, .902} unethical & \cellcolor[rgb]{ .608, .761, .902} inedible & \cellcolor[rgb]{ .608, .761, .902} bad & \cellcolor[rgb]{ .741, .843, .933} not \\
\hline
25 & \cellcolor[rgb]{ .867, .922, .969} overcharged & \cellcolor[rgb]{ .608, .761, .902} unprofessional & \cellcolor[rgb]{ .957, .69, .518} mess & \cellcolor[rgb]{ .957, .69, .518} court \\
\hline
26 & \cellcolor[rgb]{ .608, .761, .902} outdated & \cellcolor[rgb]{ .608, .761, .902} stupid & \cellcolor[rgb]{ .973, .796, .678} emails & \cellcolor[rgb]{ .867, .922, .969} said \\
\hline
27 & \cellcolor[rgb]{ .608, .761, .902} lackluster & \cellcolor[rgb]{ .608, .761, .902} incorrect & \cellcolor[rgb]{ .608, .761, .902} needless & \cellcolor[rgb]{ .973, .796, .678} minutes \\
\hline
28 & \cellcolor[rgb]{ .608, .761, .902} unprofessional & \cellcolor[rgb]{ .608, .761, .902} average & \cellcolor[rgb]{ .867, .922, .969} rushing & \cellcolor[rgb]{ .957, .69, .518} mess \\
\hline
29 & \cellcolor[rgb]{ .608, .761, .902} inedible & \cellcolor[rgb]{ .741, .843, .933} oily & \cellcolor[rgb]{ .867, .922, .969} paying & \cellcolor[rgb]{ .867, .922, .969} frustrated \\
\hline
\end{tabular}%
\label{tab:pos1}%
\end{table}%

Table \ref{tab:pos2} shows the top words contributing to non-complaints (coloring with POS). We can see rows ranking from 0 to 12 are all adjective. Adjective words are dominant in the top 40 tokens. The percentage of adjective decreases when the ranking number increases from 40 to 100. The non-complaint part of model is depending more on adjective for predictions.

There are only five nouns in top 40 tokens based on the mapping. Three of these five words are mistakenly identified as noun even they are adjective or adverb, including ``personable'', ``lovely'', ``incredibly''. The other two nouns (``recipes'', ``fun'') in top 40 words are reasonable non-complaint related words. The percentage of Nouns increase when the ranking number increases from 40 to 100. Most of these nouns with token ranks from 40 to 100 are correctly mapped nouns.

% Automatic_color_explainability.xlsx _final_pos_neg_pos
\begin{table}[htbp]
\centering
\caption{Token importance by topics (colored by POS with colors defined in Table \ref{tab:tablepos} ). Words contributing to non-complaints. Rows with all dark blue (Adjective) are removed to save space (refer to Table \ref{tab:noncomp1} for those tokens).} 
\begin{tabular}{|r|l|l|l|l|}
\hline
\rowcolor[rgb]{ .675, .725, .792} \multicolumn{1}{|p{4.055em}|}{\textbf{Rank}} & \multicolumn{1}{p{8.165em}|}{\textbf{full\_data}} & \multicolumn{1}{p{8.165em}|}{\textbf{topic: Restaurant}} & \multicolumn{1}{p{8.165em}|}{\textbf{topic: Hotel}} & \multicolumn{1}{p{8.165em}|}{\textbf{topic: Beauty}} \\
\hline
13 & \cellcolor[rgb]{ .608, .761, .902} fabulous & \cellcolor[rgb]{ .608, .761, .902} best & \cellcolor[rgb]{ .867, .922, .969} enjoyed & \cellcolor[rgb]{ .867, .922, .969} affordable \\
\hline 
15 & \cellcolor[rgb]{ .608, .761, .902} flavorful & \cellcolor[rgb]{ .608, .761, .902} yummy & \cellcolor[rgb]{ .867, .922, .969} appreciated & \cellcolor[rgb]{ .867, .922, .969} professional \\
\hline 
17 & \cellcolor[rgb]{ .741, .843, .933} timely & \cellcolor[rgb]{ .608, .761, .902} great & \cellcolor[rgb]{ .957, .69, .518} personable & \cellcolor[rgb]{ .957, .69, .518} lovely \\
\hline
18 & \cellcolor[rgb]{ .741, .843, .933} wonderfully & \cellcolor[rgb]{ .608, .761, .902} awesome & \cellcolor[rgb]{ .741, .843, .933} lovely & \cellcolor[rgb]{ .741, .843, .933} clean \\
\hline 
20 & \cellcolor[rgb]{ .867, .922, .969} entertaining & \cellcolor[rgb]{ .608, .761, .902} perfect & \cellcolor[rgb]{ .608, .761, .902} grateful & \cellcolor[rgb]{ .608, .761, .902} knowledgeable \\
\hline 
22 & \cellcolor[rgb]{ .608, .761, .902} great & \cellcolor[rgb]{ .741, .843, .933} wonderfully & \cellcolor[rgb]{ .741, .843, .933} friendly & \cellcolor[rgb]{ .741, .843, .933} love \\
\hline
23 & \cellcolor[rgb]{ .608, .761, .902} meticulous & \cellcolor[rgb]{ .741, .843, .933} timely & \cellcolor[rgb]{ .608, .761, .902} thorough & \cellcolor[rgb]{ .608, .761, .902} appreciated \\
\hline 
26 & \cellcolor[rgb]{ .608, .761, .902} unbelievable & \cellcolor[rgb]{ .608, .761, .902} respectful & \cellcolor[rgb]{ .741, .843, .933} beautifully & \cellcolor[rgb]{ .741, .843, .933} helpful \\
\hline 
28 & \cellcolor[rgb]{ .957, .69, .518} personable & \cellcolor[rgb]{ .608, .761, .902} elegant & \cellcolor[rgb]{ .867, .922, .969} love & \cellcolor[rgb]{ .867, .922, .969} friendly \\
\hline
29 & \cellcolor[rgb]{ .608, .761, .902} affordable & \cellcolor[rgb]{ .973, .796, .678} recipes & \cellcolor[rgb]{ .608, .761, .902} helpful & \cellcolor[rgb]{ .608, .761, .902} thank \\
\hline 
31 & \cellcolor[rgb]{ .608, .761, .902} respectful & \cellcolor[rgb]{ .867, .922, .969} appreciated & \cellcolor[rgb]{ .867, .922, .969} appreciate & \cellcolor[rgb]{ .867, .922, .969} honest \\
\hline
32 & \cellcolor[rgb]{ .608, .761, .902} terrific & \cellcolor[rgb]{ .608, .761, .902} romantic & \cellcolor[rgb]{ .741, .843, .933} incredibly & \cellcolor[rgb]{ .741, .843, .933} fun \\
\hline 
34 & \cellcolor[rgb]{ .867, .922, .969} appreciated & \cellcolor[rgb]{ .608, .761, .902} unbelievable & \cellcolor[rgb]{ .957, .69, .518} fun & \cellcolor[rgb]{ .957, .69, .518} incredibly \\
\hline
35 & \cellcolor[rgb]{ .608, .761, .902} romantic & \cellcolor[rgb]{ .608, .761, .902} terrific & \cellcolor[rgb]{ .867, .922, .969} recommend & \cellcolor[rgb]{ .867, .922, .969} impeccable \\
\hline 
38 & \cellcolor[rgb]{ .741, .843, .933} lovely & \cellcolor[rgb]{ .608, .761, .902} nicest & \cellcolor[rgb]{ .741, .843, .933} perfectly & \cellcolor[rgb]{ .741, .843, .933} efficient \\
\hline
39 & \cellcolor[rgb]{ .608, .761, .902} hospitable & \cellcolor[rgb]{ .867, .922, .969} love & \cellcolor[rgb]{ .608, .761, .902} honest & \cellcolor[rgb]{ .608, .761, .902} appreciate \\
\hline 
\end{tabular}%
\label{tab:pos2}%
\end{table}%

\subsubsection{Topic unique token importance}

Table \ref{tab:uniq} shows the unique tokens in each topic that are among the top 40 explainability words. For the complaint tokens, the Restaurant topic seems to be more different than the other two topics. It has a higher number of unique tokens, including many Restaurant related tokens ``tasteless'', ``rotten'', ``undercooked'' etc. The Hotel topic has the smallest number of unique words with low ranking number. The first unique token in the Hotel topic has a ranking number of 35. 

For the noncomplaint tokens, the conclusion is similar. The Restaurant is the most different than the other two topics. It has a higher number of unique tokens, including many Restaurant related words ``tasty'', ``yummy'', ``flavorful'' etc. The Beauty topic has the smallest number of unique tokens with low ranking number. The first unique token in Beauty topic has a ranking number of 25. 

% Table generated by Excel2LaTeX from sheet 'final_pos' LRP_by_topics_summary(remove 3 star).xlsx
\begin{table}[htbp]
\centering
\caption{Top explainability words that are unique to each topic.}
\begin{tabular}{|l|r|r|r|r|r|}
\hline
\rowcolor[rgb]{ .675, .725, .792} \multicolumn{3}{|p{21.145em}|}{\textbf{Unique Complaint words in a topic}} & \multicolumn{3}{p{20.995em}|}{\textbf{Unique Non-complaint words in a topic}} \\
\hline
\rowcolor[rgb]{ .675, .725, .792} \multicolumn{1}{|p{7.5em}|}{\textbf{only\_topic\_1}} & \multicolumn{1}{p{6.5em}|}{\textbf{only\_topic\_2}} & \multicolumn{1}{p{7.145em}|}{\textbf{only\_topic\_3}} & \multicolumn{1}{p{7.355em}|}{\textbf{only\_topic\_1}} & \multicolumn{1}{p{6.855em}|}{\textbf{only\_topic\_2}} & \multicolumn{1}{p{6.785em}|}{\textbf{only\_topic\_3}} \\
\hline
0.incomplete & \multicolumn{1}{l|}{\cellcolor[rgb]{ .851, .851, .851} 35.reschedule} & \multicolumn{1}{l|}{5.dishonest} & \multicolumn{1}{l|}{\cellcolor[rgb]{ .851, .851, .851} 2.superb} & \multicolumn{1}{l|}{11.meticulous} & \multicolumn{1}{l|}{\cellcolor[rgb]{ .851, .851, .851} 25.adventure} \\
\hline
1.miserable & \multicolumn{1}{l|}{\cellcolor[rgb]{ .851, .851, .851} 38.ignored} & \multicolumn{1}{l|}{8.outdated} & \multicolumn{1}{l|}{\cellcolor[rgb]{ .851, .851, .851} 3.thoughtful} & \multicolumn{1}{l|}{12.talented} & \cellcolor[rgb]{ .851, .851, .851} \\
\hline
2.tasteless & \cellcolor[rgb]{ .851, .851, .851} & \multicolumn{1}{l|}{17.unhelpful} & \multicolumn{1}{l|}{\cellcolor[rgb]{ .851, .851, .851} 4.tasty} & \multicolumn{1}{l|}{23.thorough} & \cellcolor[rgb]{ .851, .851, .851} \\
\hline
4.embarrassing & \cellcolor[rgb]{ .851, .851, .851} & \multicolumn{1}{l|}{18.lied} & \multicolumn{1}{l|}{\cellcolor[rgb]{ .851, .851, .851} 5.gracious} & \multicolumn{1}{l|}{37.gentle} & \cellcolor[rgb]{ .851, .851, .851} \\
\hline
9.horrendous & \cellcolor[rgb]{ .851, .851, .851} & \multicolumn{1}{l|}{25.court} & \multicolumn{1}{l|}{\cellcolor[rgb]{ .851, .851, .851} 8.def} & & \cellcolor[rgb]{ .851, .851, .851} \\
\hline
10.rotten & \cellcolor[rgb]{ .851, .851, .851} & \multicolumn{1}{l|}{39.complained} & \multicolumn{1}{l|}{\cellcolor[rgb]{ .851, .851, .851} 15.yummy} & & \cellcolor[rgb]{ .851, .851, .851} \\
\hline
11.incompetent & \cellcolor[rgb]{ .851, .851, .851} & & \multicolumn{1}{l|}{\cellcolor[rgb]{ .851, .851, .851} 16.flavorful} & & \cellcolor[rgb]{ .851, .851, .851} \\
\hline
12.pathetic & \cellcolor[rgb]{ .851, .851, .851} & & \multicolumn{1}{l|}{\cellcolor[rgb]{ .851, .851, .851} 21.greatest} & & \cellcolor[rgb]{ .851, .851, .851} \\
\hline
13.horribly & \cellcolor[rgb]{ .851, .851, .851} & & \multicolumn{1}{l|}{\cellcolor[rgb]{ .851, .851, .851} 22.wonderfully} & & \cellcolor[rgb]{ .851, .851, .851} \\
\hline
15.racist & \cellcolor[rgb]{ .851, .851, .851} & & \multicolumn{1}{l|}{\cellcolor[rgb]{ .851, .851, .851} 23.timely} & & \cellcolor[rgb]{ .851, .851, .851} \\
\hline
17.undercooked & \cellcolor[rgb]{ .851, .851, .851} & & \multicolumn{1}{l|}{\cellcolor[rgb]{ .851, .851, .851} 24.refreshing} & & \cellcolor[rgb]{ .851, .851, .851} \\
\hline
19.soggy & \cellcolor[rgb]{ .851, .851, .851} & & \multicolumn{1}{l|}{\cellcolor[rgb]{ .851, .851, .851} 25.thankful} & & \cellcolor[rgb]{ .851, .851, .851} \\
\hline
22.runny & \cellcolor[rgb]{ .851, .851, .851} & & \multicolumn{1}{l|}{\cellcolor[rgb]{ .851, .851, .851} 26.respectful} & & \cellcolor[rgb]{ .851, .851, .851} \\
\hline
23.lackluster & \cellcolor[rgb]{ .851, .851, .851} & & \multicolumn{1}{l|}{\cellcolor[rgb]{ .851, .851, .851} 27.hospitable} & & \cellcolor[rgb]{ .851, .851, .851} \\
\hline
24.inedible & \cellcolor[rgb]{ .851, .851, .851} & & \multicolumn{1}{l|}{\cellcolor[rgb]{ .851, .851, .851} 28.elegant} & & \cellcolor[rgb]{ .851, .851, .851} \\
\hline
26.stupid & \cellcolor[rgb]{ .851, .851, .851} & & \multicolumn{1}{l|}{\cellcolor[rgb]{ .851, .851, .851} 29.recipes} & & \cellcolor[rgb]{ .851, .851, .851} \\
\hline
27.incorrect & \cellcolor[rgb]{ .851, .851, .851} & & \multicolumn{1}{l|}{\cellcolor[rgb]{ .851, .851, .851} 30.impressive} & & \cellcolor[rgb]{ .851, .851, .851} \\
\hline
28.average & \cellcolor[rgb]{ .851, .851, .851} & & \multicolumn{1}{l|}{\cellcolor[rgb]{ .851, .851, .851} 32.romantic} & & \cellcolor[rgb]{ .851, .851, .851} \\
\hline
29.oily & \cellcolor[rgb]{ .851, .851, .851} & & \multicolumn{1}{l|}{\cellcolor[rgb]{ .851, .851, .851} 34.unbelievable} & & \cellcolor[rgb]{ .851, .851, .851} \\
\hline
30.mediocre & \cellcolor[rgb]{ .851, .851, .851} & & \multicolumn{1}{l|}{\cellcolor[rgb]{ .851, .851, .851} 35.terrific} & & \cellcolor[rgb]{ .851, .851, .851} \\
\hline
34.unfortunate & \cellcolor[rgb]{ .851, .851, .851} & & \multicolumn{1}{l|}{\cellcolor[rgb]{ .851, .851, .851} 37.hearty} & & \cellcolor[rgb]{ .851, .851, .851} \\
\hline
36.lame & \cellcolor[rgb]{ .851, .851, .851} & & \multicolumn{1}{l|}{\cellcolor[rgb]{ .851, .851, .851} 38.nicest} & & \cellcolor[rgb]{ .851, .851, .851} \\
\hline
37.angry & \cellcolor[rgb]{ .851, .851, .851} & & \cellcolor[rgb]{ .851, .851, .851} & & \cellcolor[rgb]{ .851, .851, .851} \\
\hline
38.bummed & \cellcolor[rgb]{ .851, .851, .851} & & \cellcolor[rgb]{ .851, .851, .851} & & \cellcolor[rgb]{ .851, .851, .851} \\
\hline
39.mushy & \cellcolor[rgb]{ .851, .851, .851} & & \cellcolor[rgb]{ .851, .851, .851} & & \cellcolor[rgb]{ .851, .851, .851} \\
\hline
\end{tabular}%
\label{tab:uniq}%
\end{table}%

\subsection{Gender Bias}

Explainability by segmentation can help to find potential gender related bias. There are some top words with potential gender related bias. Table \ref{tab:1modelgender} shows the potential gender related words. These words are selected using the same methods in Table \ref{tab:embedding3}. First embedding similarities of each word are calculated relative to male/female group of words. The group of male words used to calculate the embedding similarities are ['man' , 'men' , 'he' , 'his', 'sir', 'gentleman']. The group of female words used to calculate the embedding similarities are = ['woman', 'women', 'she', 'her', 'madam', 'lady']. Then the difference of the two similarities are calculated as measurement of potential male/female words. An absolute difference of similarity larger than 0.1 is considered significant in our analysis. 

Most of the potential gender words are male words in Table \ref{tab:1modelgender}. Overall, the percentage of potential gender words are small and they are not focusing on the top important words. Most of these gender related words are only slightly gender words (the similarity difference is only between 0.1 and 0.2). There are only a few strong gender related words (similarity difference greater than 0.2), including ``personable'', ``considerate'', ``courteous'', ``lovely''. Therefore there is not much concern regarding to the gender bias in the model. Only non-complaint words are shown in the table since there is significantly less gender related words in the complaint tokens. 
% Table generated by Excel2LaTeX from sheet '_final_pos_neg_pos' LRP_by_topics_summary(remove 3 star)
\begin{table}[htbp]
\centering
\caption{Words with potential gender bias (male words colored with blue and female words colored with orange). Full model trained with all topics together. }
\begin{tabular}{|l|l|r|r|}
\hline
\rowcolor[rgb]{ .675, .725, .792} \multicolumn{1}{|p{8.945em}|}{\textbf{full\_data}} & \multicolumn{1}{p{8.945em}|}{\textbf{topic: Restaurant}} & \multicolumn{1}{p{8.945em}|}{\textbf{topic: Hotel}} & \multicolumn{1}{p{8.945em}|}{\textbf{topic: Beauty}} \\
\hline
\rowcolor[rgb]{ .741, .843, .933} 3.thoughtful & 3.thoughtful & \multicolumn{1}{l|}{\cellcolor[rgb]{ .973, .796, .678} 2.beautiful} & \multicolumn{1}{l|}{\cellcolor[rgb]{ .973, .796, .678} 2.wonderful} \\
\hline
\rowcolor[rgb]{ .741, .843, .933} 4.gracious & 5.gracious & \multicolumn{1}{l|}{\cellcolor[rgb]{ .608, .761, .902} 17.personable} & \multicolumn{1}{l|}{\cellcolor[rgb]{ .608, .761, .902} 17.lovely} \\
\hline
\rowcolor[rgb]{ .973, .796, .678} 8.beautiful & \cellcolor[rgb]{ .741, .843, .933} 9.informative & \multicolumn{1}{l|}{\cellcolor[rgb]{ .741, .843, .933} 21.knowledgeable} & \multicolumn{1}{l|}{\cellcolor[rgb]{ .741, .843, .933} 21.favorite} \\
\hline
\rowcolor[rgb]{ .741, .843, .933} 9.informative & \cellcolor[rgb]{ .973, .796, .678} 14.beautiful & \multicolumn{1}{l|}{24.attentive} & \multicolumn{1}{l|}{24.outstanding} \\
\hline
\rowcolor[rgb]{ .741, .843, .933} 27.enjoyable & 19.enjoyable & \multicolumn{1}{l|}{39.honest} & \multicolumn{1}{l|}{39.appreciate} \\
\hline
\rowcolor[rgb]{ .608, .761, .902} 28.personable & \cellcolor[rgb]{ .741, .843, .933} 26.respectful & \multicolumn{1}{l|}{\cellcolor[rgb]{ .741, .843, .933} 46.decent} & \multicolumn{1}{l|}{\cellcolor[rgb]{ .741, .843, .933} 46.gift} \\
\hline
\rowcolor[rgb]{ .741, .843, .933} 31.respectful & 27.hospitable & \multicolumn{1}{l|}{\cellcolor[rgb]{ .973, .796, .678} 54.beauty} & \multicolumn{1}{l|}{\cellcolor[rgb]{ .973, .796, .678} 54.truly} \\
\hline
\rowcolor[rgb]{ .741, .843, .933} 39.hospitable & \cellcolor[rgb]{ .973, .796, .678} 29.recipes & \cellcolor[rgb]{ 1, 1, 1} & \cellcolor[rgb]{ 1, 1, 1} \\
\hline
\rowcolor[rgb]{ .741, .843, .933} 40.knowledgeable & 30.impressive & \cellcolor[rgb]{ 1, 1, 1} & \cellcolor[rgb]{ 1, 1, 1} \\
\hline
\rowcolor[rgb]{ .741, .843, .933} 41.impressive & 38.nicest & \cellcolor[rgb]{ 1, 1, 1} & \cellcolor[rgb]{ 1, 1, 1} \\
\hline
\rowcolor[rgb]{ .608, .761, .902} 44.considerate & 45.personable & \cellcolor[rgb]{ 1, 1, 1} & \cellcolor[rgb]{ 1, 1, 1} \\
\hline
\rowcolor[rgb]{ .741, .843, .933} 57.delightful & 51.delightful & \cellcolor[rgb]{ 1, 1, 1} & \cellcolor[rgb]{ 1, 1, 1} \\
\hline
\rowcolor[rgb]{ .741, .843, .933} 58.nicest & 56.knowledgeable & \cellcolor[rgb]{ 1, 1, 1} & \cellcolor[rgb]{ 1, 1, 1} \\
\hline
\rowcolor[rgb]{ .741, .843, .933} 68.attentive & 62.attentive & \cellcolor[rgb]{ 1, 1, 1} & \cellcolor[rgb]{ 1, 1, 1} \\
\hline
\rowcolor[rgb]{ .741, .843, .933} 82.charming & 78.charming & \cellcolor[rgb]{ 1, 1, 1} & \cellcolor[rgb]{ 1, 1, 1} \\
\hline
\rowcolor[rgb]{ .608, .761, .902} 85.courteous & 88.courteous & \cellcolor[rgb]{ 1, 1, 1} & \cellcolor[rgb]{ 1, 1, 1} \\
\hline
\rowcolor[rgb]{ .741, .843, .933} 96.honest & 91.impeccable & \cellcolor[rgb]{ 1, 1, 1} & \cellcolor[rgb]{ 1, 1, 1} \\
\hline
\end{tabular}%
\label{tab:1modelgender}%
\end{table}%

% Table generated by Excel2LaTeX from sheet '_final_pos_neg_pos' LRP_by_topics_summary(remove 3 star)
\begin{table}[htbp]
\centering
\caption{Words with potential gender bias (male words colored with blue and female words colored with orange). Separate models trained with each topics separately.}% Table generated by Excel2LaTeX from sheet 'final_pos_4models}
\begin{tabular}{|r|r|l|l|}
\hline
\rowcolor[rgb]{ .675, .725, .792} \multicolumn{1}{|p{9.055em}|}{\textbf{full\_data}} & \multicolumn{1}{p{9.055em}|}{\textbf{topic: Restaurant}} & \multicolumn{1}{p{9.055em}|}{\textbf{topic: Hotel}} & \multicolumn{1}{p{8.22em}|}{\textbf{topic: Beauty}} \\
\hline
\rowcolor[rgb]{ .741, .843, .933} \multicolumn{1}{|l|}{3.thoughtful} & \multicolumn{1}{l|}{\cellcolor[rgb]{ .973, .796, .678} 8.beautiful} & \cellcolor[rgb]{ .973, .796, .678} 5.beautiful & 0.roomy \\
\hline
\rowcolor[rgb]{ .741, .843, .933} \multicolumn{1}{|l|}{4.gracious} & \multicolumn{1}{l|}{13.gracious} & 6.enjoyable & 1.thoughtful \\
\hline
\rowcolor[rgb]{ .973, .796, .678} \multicolumn{1}{|l|}{8.beautiful} & \multicolumn{1}{l|}{\cellcolor[rgb]{ .741, .843, .933} 18.informative} & \cellcolor[rgb]{ .741, .843, .933} 16.attentive & \cellcolor[rgb]{ .741, .843, .933} 2.informative \\
\hline
\rowcolor[rgb]{ .741, .843, .933} \multicolumn{1}{|l|}{9.informative} & \multicolumn{1}{l|}{21.thoughtful} & 23.knowledgeable & \cellcolor[rgb]{ .608, .761, .902} 3.personable \\
\hline
\rowcolor[rgb]{ .741, .843, .933} \multicolumn{1}{|l|}{27.enjoyable} & \multicolumn{1}{l|}{24.hospitable} & \cellcolor[rgb]{ .608, .761, .902} 26.personable & 5.enjoyable \\
\hline
\rowcolor[rgb]{ .608, .761, .902} \multicolumn{1}{|l|}{28.personable} & \multicolumn{1}{l|}{\cellcolor[rgb]{ .741, .843, .933} 26.impressive} & \cellcolor[rgb]{ .741, .843, .933} 27.respectful & \cellcolor[rgb]{ .741, .843, .933} 9.respectful \\
\hline
\rowcolor[rgb]{ .741, .843, .933} \multicolumn{1}{|l|}{31.respectful} & \multicolumn{1}{l|}{\cellcolor[rgb]{ .608, .761, .902} 28.personable} & \cellcolor[rgb]{ .608, .761, .902} 36.considerate & \cellcolor[rgb]{ .608, .761, .902} 13.courteous \\
\hline
\rowcolor[rgb]{ .741, .843, .933} \multicolumn{1}{|l|}{39.hospitable} & \multicolumn{1}{l|}{\cellcolor[rgb]{ .973, .796, .678} 31.recipes} & 38.thoughtful & \cellcolor[rgb]{ .608, .761, .902} 16.polite \\
\hline
\rowcolor[rgb]{ .741, .843, .933} \multicolumn{1}{|l|}{40.knowledgeable} & \multicolumn{1}{l|}{47.delightful} & 43.informative & 20.hospitable \\
\hline
\rowcolor[rgb]{ .741, .843, .933} \multicolumn{1}{|l|}{41.impressive} & \multicolumn{1}{l|}{48.enjoyable} & 46.impressive & 21.gracious \\
\hline
\rowcolor[rgb]{ .608, .761, .902} \multicolumn{1}{|l|}{44.considerate} & \multicolumn{1}{l|}{\cellcolor[rgb]{ .741, .843, .933} 56.nicest} & \cellcolor[rgb]{ .973, .796, .678} 47.beauty & \cellcolor[rgb]{ .741, .843, .933} 23.knowledgeable \\
\hline
\rowcolor[rgb]{ .741, .843, .933} \multicolumn{1}{|l|}{57.delightful} & \multicolumn{1}{l|}{68.respectful} & 48.decent & \cellcolor[rgb]{ .973, .796, .678} 27.beautiful \\
\hline
\rowcolor[rgb]{ .741, .843, .933} \multicolumn{1}{|l|}{58.nicest} & \multicolumn{1}{l|}{\cellcolor[rgb]{ .973, .796, .678} 97.joy} & \cellcolor[rgb]{ .973, .796, .678} \textbf{50.tina} & 31.honest \\
\hline
\rowcolor[rgb]{ .741, .843, .933} \multicolumn{1}{|l|}{68.attentive} & \multicolumn{1}{l|}{98.charming} & \cellcolor[rgb]{ .973, .796, .678} 57.joy & \cellcolor[rgb]{ .973, .796, .678} 37.scenic \\
\hline
\rowcolor[rgb]{ .741, .843, .933} \multicolumn{1}{|l|}{82.charming} & \cellcolor[rgb]{ 1, 1, 1} & \cellcolor[rgb]{ .957, .69, .518} 66.goddess & 38.charming \\
\hline
\rowcolor[rgb]{ .608, .761, .902} \multicolumn{1}{|l|}{85.courteous} & \cellcolor[rgb]{ 1, 1, 1} & \cellcolor[rgb]{ .973, .796, .678} \textbf{71.natalie} & 40.considerate \\
\hline
\rowcolor[rgb]{ .741, .843, .933} \multicolumn{1}{|l|}{96.honest} & \cellcolor[rgb]{ 1, 1, 1} & \cellcolor[rgb]{ .973, .796, .678} 83.mimi & 59.impressive \\
\hline
& & \cellcolor[rgb]{ .957, .69, .518} \textbf{84.jennifer} & \cellcolor[rgb]{ .741, .843, .933} 62.attentive \\
\hline
& & \cellcolor[rgb]{ .741, .843, .933} 86.charming & \cellcolor[rgb]{ .741, .843, .933} 65.cheerful \\
\hline
& & \cellcolor[rgb]{ .973, .796, .678} \textbf{87.annie} & \cellcolor[rgb]{ .741, .843, .933} 66.decent \\
\hline
& & \cellcolor[rgb]{ .741, .843, .933} 88.jan & \cellcolor[rgb]{ .741, .843, .933} 90.nicest \\
\hline
& & \cellcolor[rgb]{ .973, .796, .678} \textbf{89.grace} & \cellcolor[rgb]{ .741, .843, .933} 96.delightful \\
\hline
\end{tabular}%
\label{tab:4modelgender}%
\end{table}%

If separate models are trained for each topics, there are more gender related top tokens, as shown in Table \ref{tab:4modelgender}, especially in the Hotel and Beauty topics. There are some special gender related female words, such as ``tina'', ``natalie'', which are people’s names. This is something we want to avoid in a model. If the model is relying on some female names to make prediction, the model could be biased. The impact might be small in this model since these female name words do not have very high frequencies in the data and their ranking number is high. These results show that, in this data, fitting one large model with all topics can help to decrease the gender related bias.

\subsection{False Positive and False Negative Error Analysis}

Local explanations have lots of detailed information regarding the model explainability. However, it requires large amount of work to review all the local explainability and provide feedback. It would be more effective to review the False Positive predictions and False negative predictions, similarly as the explanations in \citep{Khanna2019black} and \citep{Han2020subset}. 

False positive predictions are caused by some positive words in the reviews. Understanding these positive words’ role in false positive predictions will be helpful to understand why the model fails. Similarly, false negative predictions are caused by some negative words in the reviews. Similarly, these negative words’ role in false negative predictions are very important.

\bgroup
%%%%%%%%%% topic: Restaurant %%%%%%%%%%%%%%% 
{\setlength{\fboxsep}{ 1.5pt} \setlength{\fboxrule}{0pt} \colorbox{white!0} 
%\usepackage{adjustbox}
%\tcbset{width=0.9\textwidth,boxrule=0pt,colback=red,arc=0pt,auto outer arc,left=0pt,right=0pt,boxsep=5pt}

\def\arraystretch{1.9}% 1 is the default, change whatever you need 
%\vspace{0.753in} 
\begin{table*}[t] 
\caption{The Restaurant topic false positive examples, highlighted based on the LRP scores. Blue tokens are contributing to the class 1 (Complaint). }
\centering 
\begin{tabular}{|p{0.5cm}|p{0.5cm}|p{0.5cm}|p{11.5cm}|p{3.0cm}|} 
%\begin{tabular}{| >{\normalsize}p{0.5cm}|>{\normalsize}p{0.5cm}|>{\normalsize}p{0.5cm}|>{\normalsize}p{0.5cm}|>{\normalsize}p{12cm}|} 
\hline
Act & Pred & Prob & Review Label = 1 (Complaint ), Label = 0 (Non Complaint ) & Comments \\
\hline

0 & 1.0 & 0.52 & \colorbox{red!68.0}{\strut yummy} scratch made food in a\colorbox{red!14.0}{\strut casual} and non\colorbox{blue!100.0}{\strut stuffy} \colorbox{blue!26.0}{\strut environment} \colorbox{blue!61.0}{\strut prices} are very reasonable and service is\colorbox{red!20.0}{\strut always} \colorbox{red!25.0}{\strut spot} on this is our\colorbox{red!23.0}{\strut italian} comfort food place and there has\colorbox{blue!50.0}{\strut not} been a single\colorbox{red!26.0}{\strut thing} i have tried that i did\colorbox{blue!50.0}{\strut not} \colorbox{blue!25.0}{\strut like} 
& The negative word `non' before `stuffy' is not used by the model. \\ \hline 
0 & 1.0 & 0.52 & the size was a little\colorbox{blue!100.0}{\strut disappointing} as\colorbox{blue!22.0}{\strut compared} to the\colorbox{blue!28.0}{\strut price} though that\colorbox{blue!24.0}{\strut smoked} meat was\colorbox{red!29.0}{\strut heavenly} \colorbox{red!57.0}{\strut good} \colorbox{red!57.0}{\strut good} \colorbox{blue!13.0}{\strut service} and covid protocols
& Size is a little “disappointing”. Very good service. \\ \hline 
0 & 1.0 & 0.81 & i stopped by\colorbox{blue!60.0}{\strut yesterday} and there was an\colorbox{blue!99.0}{\strut eviction} \colorbox{blue!23.0}{\strut notice} on the\colorbox{blue!12.0}{\strut door} and all of their\colorbox{blue!27.0}{\strut furniture} was gone stopped by for\colorbox{red!11.0}{\strut oktoberfest} \colorbox{blue!15.0}{\strut last} fall and was looking\colorbox{red!26.0}{\strut forward} to coming\colorbox{red!32.0}{\strut back} but\colorbox{blue!30.0}{\strut sadly} covid has claimed another\colorbox{blue!13.0}{\strut business} there are\colorbox{blue!89.0}{\strut not} a\colorbox{red!22.0}{\strut lot} of\colorbox{red!100.0}{\strut good} \colorbox{red!33.0}{\strut german} food\colorbox{red!37.0}{\strut options} in\colorbox{red!16.0}{\strut columbus} and now there is one\colorbox{blue!18.0}{\strut fewer} 
& It is not easy to tell from the words in the comments it's a good review. \\ \hline 
\hline
\end{tabular}
\label{tab:table1fp}
\end{table*} 
} 
In one topic (topic: Restaurant) FP example in Table \ref{tab:table1fp}, a negative word before a bad word (`stuffy') is not recognized by the model. In another example in Table \ref{tab:table1fp}, the model is misled by the `disappointing' size comment from the customer. The customer still rated good stars for the Restaurant despite the small size of food. The model is not very sure about these two predictions (probability close to 0.5). In the last example in Table \ref{tab:table1fp}, the label is a little tricky. It's a customer's comments about a closed Restaurant. The customer likes the Restaurant. However, the comments just mentioned something regarding the close of the Restaurant. It's not easy to tell from the words it's a non-complaint review.

In one topic (topic: Restaurant) FN example in Table \ref{tab:table1fn}, there are non-complaint words related to other people's yelp review before visit. This contributed to the non-complaint prediction of the FN example. In another example in Table \ref{tab:table1fn}, model is confused by `better burgers' at other place (lead to non-complaint prediction). In last example in Table \ref{tab:table1fn}, customer used comparison to show the food is cheap and low quality. Model is not smart enough to tell that the compared food is lower quality since there is no obvious bad words. 

{\setlength{\fboxsep}{ 1.5pt} \setlength{\fboxrule}{0pt} \colorbox{white!0} 
%\usepackage{adjustbox}
%\tcbset{width=0.9\textwidth,boxrule=0pt,colback=red,arc=0pt,auto outer arc,left=0pt,right=0pt,boxsep=5pt}

\def\arraystretch{1.9}% 1 is the default, change whatever you need 
%\vspace{0.753in} 
\begin{table*}[t] 
\caption{the Restaurant topic false negative examples, highlighted based on the LRP scores. Blue tokens are contributing to the class 1 (Complaint). }
\centering
\begin{tabular}{|p{0.5cm}|p{0.5cm}|p{0.5cm}|p{11.5cm}|p{3.0cm}|} 
%\begin{tabular}{| >{\normalsize}p{0.5cm}|>{\normalsize}p{0.5cm}|>{\normalsize}p{0.5cm}|>{\normalsize}p{0.5cm}|>{\normalsize}p{12cm}|} 
\hline
Act & Pred & Prob & Review Label = 1 (Complaint ), Label = 0 (Non Complaint ) & Comments \\
\hline
1 & 0.0 & 0.31 & if you have ever been to\colorbox{red!33.0}{\strut san} diego you will\colorbox{blue!15.0}{\strut know} that their mexican food is\colorbox{red!57.0}{\strut top} \colorbox{red!53.0}{\strut tier} i was\colorbox{red!15.0}{\strut browsing} \colorbox{red!23.0}{\strut yelp} one day and got\colorbox{red!64.0}{\strut pretty} \colorbox{red!51.0}{\strut excited} when i saw they had carne asada fries on the\colorbox{red!51.0}{\strut menu} and it s a\colorbox{red!22.0}{\strut new} business we were in the\colorbox{red!43.0}{\strut area} so decided to give it a\colorbox{red!43.0}{\strut try} i thought it was\colorbox{blue!19.0}{\strut weird} they only did\colorbox{blue!26.0}{\strut orders} through uberonline but i\colorbox{blue!11.0}{\strut guess} it s because they are\colorbox{blue!24.0}{\strut still} in the\colorbox{blue!13.0}{\strut works} the order took a\colorbox{red!15.0}{\strut little} \colorbox{blue!23.0}{\strut longer} than the\colorbox{blue!76.0}{\strut estimated} time but it was\colorbox{blue!57.0}{\strut not} too\colorbox{blue!44.0}{\strut bad} of a\colorbox{red!11.0}{\strut wait} \colorbox{blue!100.0}{\strut sad} to\colorbox{blue!32.0}{\strut say} the fries were just\colorbox{blue!18.0}{\strut okay} the\colorbox{red!14.0}{\strut photos} and everything\colorbox{red!15.0}{\strut looks} so much\colorbox{red!15.0}{\strut better} they were\colorbox{red!17.0}{\strut definitely} \colorbox{red!35.0}{\strut loaded} fries rather than carne asada fries 
& The positive words from other people's yelp review before visit inpacted the overall prediction.
\\ \hline 
1 & 0.0 & 0.29 & this is\colorbox{blue!40.0}{\strut low} \colorbox{blue!51.0}{\strut grade} \colorbox{red!29.0}{\strut dog} \colorbox{red!24.0}{\strut food} i have had\colorbox{red!100.0}{\strut better} \colorbox{red!88.0}{\strut burgers} at\colorbox{red!55.0}{\strut wendy} s and for\colorbox{blue!58.0}{\strut half} the\colorbox{red!13.0}{\strut price} 
& Model is confused by `better burgers' at other place. \\ \hline 

1 & 0.0 & 0.13 & stays open on\colorbox{red!13.0}{\strut tourism} \colorbox{blue!24.0}{\strut money} and i can\colorbox{blue!100.0}{\strut not} \colorbox{blue!14.0}{\strut imagine} there are any regulars the\colorbox{red!23.0}{\strut food} was so\colorbox{red!79.0}{\strut incredibly} cheap\colorbox{red!14.0}{\strut tasting} and\colorbox{blue!22.0}{\strut bland} \colorbox{red!13.0}{\strut lobster} roll was on what\colorbox{blue!11.0}{\strut tasted} \colorbox{blue!11.0}{\strut like} a cheap\colorbox{red!30.0}{\strut hot} \colorbox{red!13.0}{\strut dog} bun\colorbox{red!18.0}{\strut cocktails} taste\colorbox{blue!11.0}{\strut like} something at a\colorbox{blue!41.0}{\strut college} \colorbox{red!14.0}{\strut party} just alcohol and\colorbox{blue!30.0}{\strut overly} \colorbox{red!60.0}{\strut sweet} \colorbox{red!43.0}{\strut mixer} come for a\colorbox{red!30.0}{\strut beer} and\colorbox{red!73.0}{\strut music} if your\colorbox{red!16.0}{\strut looking} for that but stay away from everything else 
& Customer used comparison to show the food is cheap. Model cannot tell this. \\ \hline

\hline
\end{tabular}
\label{tab:table1fn}
\end{table*} 
} 

%%%%%%%%%% topic: Hotel %%%%%%%%%%%%%%% 

{\setlength{\fboxsep}{ 1.5pt} \setlength{\fboxrule}{0pt} \colorbox{white!0} 
%\usepackage{adjustbox}
%\tcbset{width=0.9\textwidth,boxrule=0pt,colback=red,arc=0pt,auto outer arc,left=0pt,right=0pt,boxsep=5pt}

\def\arraystretch{1.9}% 1 is the default, change whatever you need 
%\vspace{0.753in} 
\begin{table*}[t] 
\caption{The Hotel topic false positive examples, highlighted based on the LRP scores. Blue tokens are contributing to the class 1 (Complaint). }
\centering 
\begin{tabular}{|p{0.5cm}|p{0.5cm}|p{0.5cm}|p{11.5cm}|p{3.0cm}|} 
%\begin{tabular}{| >{\normalsize}p{0.5cm}|>{\normalsize}p{0.5cm}|>{\normalsize}p{0.5cm}|>{\normalsize}p{0.5cm}|>{\normalsize}p{12cm}|} 
\hline
Act & Pred & Prob & Review Label = 1 (Complaint ), Label = 0 (Non Complaint ) & Comments \\
\hline

0 & 1.0 & 0.93 & always\colorbox{red!77.0}{\strut love} \colorbox{red!22.0}{\strut good} \colorbox{red!12.0}{\strut nail} \colorbox{red!12.0}{\strut salon} offers online booking although place not online booking request appt yelp responded within day confirm\colorbox{blue!17.0}{\strut friend} decided try new place today\colorbox{blue!12.0}{\strut left} liking\colorbox{blue!11.0}{\strut nails} service\colorbox{blue!22.0}{\strut received} parking convenient free parking lot right behind building finding parking\colorbox{blue!11.0}{\strut wo} not issue n williams the thing bit\colorbox{blue!100.0}{\strut unprofessional} time pay kept repeating preferred cash even though accept credit cards get not prefer cash\colorbox{blue!23.0}{\strut customer} not\colorbox{red!12.0}{\strut feel} \colorbox{blue!14.0}{\strut guilt} wanting pay credit card\colorbox{blue!16.0}{\strut debit} card cash overall\colorbox{red!32.0}{\strut enjoyed} \colorbox{red!22.0}{\strut quality} service back 
& `unprofessional' is referring to payment service. The overall service is still good. \\ \hline 
0 & 1.0 & 0.9 & \colorbox{red!51.0}{\strut great} \colorbox{red!30.0}{\strut place} for\colorbox{red!17.0}{\strut walkin} situation last time i\colorbox{blue!30.0}{\strut went} was\colorbox{blue!56.0}{\strut empty} \colorbox{blue!38.0}{\strut due} to covid\colorbox{red!42.0}{\strut im} \colorbox{blue!46.0}{\strut not} sure what all they are doing with\colorbox{red!19.0}{\strut safety} during covid but i\colorbox{red!13.0}{\strut wore} a mask and did\colorbox{blue!46.0}{\strut not} \colorbox{red!26.0}{\strut feel} like things were\colorbox{blue!100.0}{\strut dirty} i did\colorbox{blue!46.0}{\strut not} \colorbox{red!26.0}{\strut feel} like i was at\colorbox{red!11.0}{\strut risk} of some sort of contamination 
& The model cannot recognize the negative expression related to `dirty'. \\ \hline 
\hline
\end{tabular}
\label{tab:table2fp}
\end{table*} 
}

The complaint words in one topic (topic: Hotel) FP example in Table \ref{tab:table2fp} are related to less important part of the service (payment system), the customer is still satisfied with the overall service. The model is not smart enough to make the distinction. It's a challenge review for the model. Another example in Table \ref{tab:table2fp} is caused by the ignorance of negative words before a complaint related word `dirty'.

{\setlength{\fboxsep}{ 1.5pt} \setlength{\fboxrule}{0pt} \colorbox{white!0} 
%\usepackage{adjustbox}
%\tcbset{width=0.9\textwidth,boxrule=0pt,colback=red,arc=0pt,auto outer arc,left=0pt,right=0pt,boxsep=5pt}

\def\arraystretch{1.9}% 1 is the default, change whatever you need 
%\vspace{0.753in} 
\begin{table*}[t] 
\caption{The Hotel topic false negative examples, highlighted based on the LRP scores. Blue tokens are contributing to the class 1 (Complaint). }
\centering
\begin{tabular}{|p{0.5cm}|p{0.5cm}|p{0.5cm}|p{11.5cm}|p{3.0cm}|} 
%\begin{tabular}{| >{\normalsize}p{0.5cm}|>{\normalsize}p{0.5cm}|>{\normalsize}p{0.5cm}|>{\normalsize}p{0.5cm}|>{\normalsize}p{12cm}|} 
\hline
Act & Pred & Prob & Review Label = 1 (Complaint ), Label = 0 (Non Complaint ) & Comments \\
\hline
1 & 0.0 & 0.18 & my first to suzy q was\colorbox{red!32.0}{\strut great} \colorbox{red!16.0}{\strut jordan} did an\colorbox{red!100.0}{\strut amazing} \colorbox{red!31.0}{\strut job} took her time and made me feel like\colorbox{red!12.0}{\strut i} was getting my\colorbox{blue!26.0}{\strut money} s worth today\colorbox{red!12.0}{\strut i} went and\colorbox{red!12.0}{\strut i} failed to get the young ladies name who did my pedicure she was young and seemed to be in hurry my\colorbox{blue!16.0}{\strut polish} is\colorbox{blue!14.0}{\strut not} \colorbox{blue!34.0}{\strut totally} \colorbox{blue!11.0}{\strut even} and my legs\colorbox{blue!11.0}{\strut still} have exfoliate and lotion\colorbox{blue!14.0}{\strut not} rubbed in\colorbox{red!12.0}{\strut i} will give suzy q one my try 
& Last service with Jordan was `amazing'. Today's service by another lady is not.
\\ \hline 
1 & 0.0 & 0.5 & their\colorbox{red!23.0}{\strut prices} are\colorbox{blue!15.0}{\strut extremely} \colorbox{blue!31.0}{\strut high} i was\colorbox{blue!100.0}{\strut charged} for an brow thread service\colorbox{red!32.0}{\strut plenty} of other options around the hillsborobeaverton\colorbox{red!45.0}{\strut area} that do a much\colorbox{red!29.0}{\strut better} \colorbox{red!51.0}{\strut job} for\colorbox{blue!14.0}{\strut fractions} of the price 
& Model is confused by `much better job' in other places. \\ \hline 
\hline
\end{tabular}
\label{tab:table2fn}
\end{table*} 
} 

The nice words in one topic (topic: Hotel) FN example in Table \ref{tab:table2fn} are related to a previous service. Today's service by another employee is very bad. The model cannot tell the differences between the two services and got confused. Another example in Table \ref{tab:table2fn} is similar. The customer mentioned some nice words about other places and confused the model. 
%%%%%%%%%% topic: Beauty %%%%%%%%%%%%%%% 

{\setlength{\fboxsep}{ 1.5pt} \setlength{\fboxrule}{0pt} \colorbox{white!0} 
%\usepackage{adjustbox}
%\tcbset{width=0.9\textwidth,boxrule=0pt,colback=red,arc=0pt,auto outer arc,left=0pt,right=0pt,boxsep=5pt}

\def\arraystretch{1.9}% 1 is the default, change whatever you need 
%\vspace{0.753in} 
\begin{table*}[t] 
\caption{The Beauty topic false positive examples, highlighted based on the LRP scores. Blue tokens are contributing to the class 1 (Complaint). }
\centering 
\begin{tabular}{|p{0.5cm}|p{0.5cm}|p{0.5cm}|p{11.5cm}|p{3.0cm}|} 
%\begin{tabular}{| >{\normalsize}p{0.5cm}|>{\normalsize}p{0.5cm}|>{\normalsize}p{0.5cm}|>{\normalsize}p{0.5cm}|>{\normalsize}p{12cm}|} 
\hline
Act & Pred & Prob & Review Label = 1 (Complaint ), Label = 0 (Non Complaint ) & Comments \\
\hline

0 & 1.0 & 0.95 & i stayed here from january\colorbox{red!41.0}{\strut th} to\colorbox{red!41.0}{\strut th} here was very\colorbox{blue!14.0}{\strut cheap} and\colorbox{blue!70.0}{\strut budget} hoteli was at disney theme\colorbox{red!47.0}{\strut parks} from am to\colorbox{blue!14.0}{\strut pmi} used this\colorbox{blue!34.0}{\strut hotel} in order to sleepin my opinion it is a\colorbox{blue!78.0}{\strut waste} of\colorbox{blue!100.0}{\strut money} to stay an\colorbox{blue!11.0}{\strut expensive} \colorbox{blue!34.0}{\strut hotel} for only\colorbox{blue!37.0}{\strut sleep} people were very nicethey\colorbox{blue!15.0}{\strut tried} to help\colorbox{blue!30.0}{\strut me} i did\colorbox{blue!32.0}{\strut not} know a hidden chargersvp feethe\colorbox{blue!34.0}{\strut hotel} \colorbox{blue!16.0}{\strut charged} \colorbox{blue!15.0}{\strut per} \colorbox{blue!37.0}{\strut night} i used\colorbox{red!26.0}{\strut shuttle} \colorbox{red!20.0}{\strut buses} to disney worldwhen i stayed therei used at am on my way to the theme parksthen i used at\colorbox{blue!78.0}{\strut pm} on my way to the\colorbox{blue!34.0}{\strut hotel} the\colorbox{blue!34.0}{\strut hotel} did\colorbox{blue!32.0}{\strut not} serve breakfastno restaurantthe\colorbox{red!23.0}{\strut pizza} hut was\colorbox{blue!44.0}{\strut closed} ...... 
& A challenge one to label. More complaint words than non-complaint words, even it's not a complaint. \\ \hline 
0 & 1.0 & 0.86 & \colorbox{red!41.0}{\strut excellent} service\colorbox{blue!14.0}{\strut never} a problemalways\colorbox{red!20.0}{\strut greeted} by a\colorbox{red!68.0}{\strut great} \colorbox{red!18.0}{\strut driver} be\colorbox{red!39.0}{\strut grateful} and stop tossing\colorbox{blue!49.0}{\strut hate} \colorbox{blue!90.0}{\strut remarks} because you are a\colorbox{blue!100.0}{\strut miserable} person 
& Model did not recognize the word `stop' before the bad words `hate remarks'. \\ \hline 
\hline
\end{tabular}
\label{tab:table3fp}
\end{table*} 
}

In one topic (topic: Beauty) FP example in Table \ref{tab:table3fp}, there are more complaint words than non-complaint words, even it's not a complaint. It's a challenge one to predict. Especially the expression `waste of money to stay an expensive hotel' can be easily interpreted as a complaint when these words are looked separately. Another example in Table \ref{tab:table3fp} is caused by the ignorance of word `stop' before a complaint related word `hate remarks'. This belongs to the same category of ignorance of negative words. This is a very common pattern in FP examples.

{\setlength{\fboxsep}{ 1.5pt} \setlength{\fboxrule}{0pt} \colorbox{white!0} 
%\usepackage{adjustbox}
%\tcbset{width=0.9\textwidth,boxrule=0pt,colback=red,arc=0pt,auto outer arc,left=0pt,right=0pt,boxsep=5pt}

\def\arraystretch{1.9}% 1 is the default, change whatever you need 
%\vspace{0.753in} 
\begin{table*}[t] 
\caption{The Beauty topic false negative examples, highlighted based on the LRP scores. Blue tokens are contributing to the class 1 (Complaint). }
\centering
\begin{tabular}{|p{0.5cm}|p{0.5cm}|p{0.5cm}|p{11.5cm}|p{3.0cm}|} 
%\begin{tabular}{| >{\normalsize}p{0.5cm}|>{\normalsize}p{0.5cm}|>{\normalsize}p{0.5cm}|>{\normalsize}p{0.5cm}|>{\normalsize}p{12cm}|} 
\hline
Act & Pred & Prob & Review Label = 1 (Complaint ), Label = 0 (Non Complaint ) & Comments \\
\hline
1 & 0.0 & 0.08 & getting a little\colorbox{red!16.0}{\strut long} in the tooth time to\colorbox{blue!25.0}{\strut remodel} the\colorbox{blue!44.0}{\strut bathroom} \colorbox{red!24.0}{\strut wall} was coming apart\colorbox{blue!30.0}{\strut no} \colorbox{blue!18.0}{\strut usb} \colorbox{red!51.0}{\strut ports} at all and the outlets on all the\colorbox{blue!13.0}{\strut lights} had so much play that\colorbox{blue!30.0}{\strut no} twoprong\colorbox{blue!21.0}{\strut plugs} \colorbox{blue!50.0}{\strut would} connect had\colorbox{red!18.0}{\strut one} \colorbox{red!24.0}{\strut wall} \colorbox{red!27.0}{\strut outlet} that was\colorbox{red!40.0}{\strut helpful} next to the\colorbox{blue!21.0}{\strut bed} \colorbox{red!25.0}{\strut staff} was\colorbox{red!52.0}{\strut nice} \colorbox{red!27.0}{\strut location} has\colorbox{red!48.0}{\strut plenty} of\colorbox{red!100.0}{\strut casual} \colorbox{red!24.0}{\strut food} \colorbox{red!23.0}{\strut nearby} and\colorbox{red!11.0}{\strut convenient} to 
& Many strong nice words about staff and location, even it's a complaint review.
\\ \hline 
1 & 0.0 & 0.07 & pros connected to\colorbox{red!19.0}{\strut convention} \colorbox{red!23.0}{\strut center} by a\colorbox{red!25.0}{\strut walkway} \colorbox{red!64.0}{\strut clean} \colorbox{red!18.0}{\strut cons} \colorbox{red!57.0}{\strut very} noisy\colorbox{red!43.0}{\strut thanks} to\colorbox{red!62.0}{\strut music} from the\colorbox{red!48.0}{\strut club} below the\colorbox{red!62.0}{\strut music} \colorbox{red!21.0}{\strut continued} \colorbox{blue!43.0}{\strut till} am and i was\colorbox{blue!100.0}{\strut completely} \colorbox{blue!56.0}{\strut unable} to\colorbox{blue!21.0}{\strut rest} at\colorbox{blue!14.0}{\strut night} 
& Ironic expression `thanks to music from the club' is hard for the model to understand.
\\ \hline 
\hline
\end{tabular}
\label{tab:table3fn}
\end{table*} 
}

\egroup

In one topic (topic: Beauty) FN example in Table \ref{tab:table3fn}, there are many strong nice words about staff and location in a complaint review which makes the task of prediction very difficult. Another example in Table \ref{tab:table3fn} is caused by the ironic expression `thanks to music from the club' which is hard for the model to understand.

After reviewing these FP and FN examples, we have the following observations. 
\begin{enumerate}
\item Some of the FP/FN are caused by challenge observations, where there are more complaint words in a non-complaint review or there are more non-complaint words in complaint review. Even people can got confused by these reviews. 
\item There are other situations that part of the complaint/non-complaint review is talking about nice/bad things not directly or closely related to the main subject of the review. People can easily tell from the context of the review that it's a complaint or not. However, these reviews are challenging to the model which mostly rely on the individual word to do the prediction. 

\item In FP examples, ignorance of negative words is a very common pattern. Negative words by themselves are usually treated as complaint words by the model. This is problematic if the negative words are used before some complaint words. They should offset the complaint words. However, the model usually cannot achieve this goal and both the negative words and complaint words are contributing to the FP error. 

\end{enumerate}

These challenges are expected for an NLP model when the data size is limited and the model is not very huge. After reviewing these FP/FN examples, we are more comfortable with the model performance even though they have some limitations. No obvious model mistakes of the models have been observed in these FP/FN analysis. 

Reviewing the FP/FN by different topics gives us more representative samples. It also tells different stories in different topics. For example, `cheap' food in Restaurant usually is a bad thing especially when people have high expectation. However, some people may enjoy `cheap' Hotel since they believe it's a `waste of money to stay an expensive hotel'.

\section{Conclusion}

In this paper, we discussed a method to analyze the NLP model explainability by segments.
\begin{enumerate}
\item Analyzing the model explainability by topics provides a lot more information regarding to an NLP classification model. Especially when this is combined with embedding similarity analysis and sentiment analysis of the top tokens. 
\item In this paper, we also discussed how to use the Part of Speech information to analyze the top explainability words. It shows that most of the words contributing to the model are adjectives. This information might be used to improve the model performance in the future.

\item Some analysis of gender bias is done based on embedding similarity of the top tokens to the gender words. It provides a useful tool to test the NLP model bias. 

\item FP/FN Error analysis provides more insights on what caused the model to be wrong. Doing the error analysis at segment level help people to understand the challenges of the NLP models. 

\end{enumerate}

\section{Acknowledgment}
We thank Harsh Singhal, Jie Chen, Vijayan Nair, Tarun Joshi, Xin Yan, and Ye Yu for insightful discussion. We thank corporate risk - model risk at Wells Fargo for support.
The views expressed in the paper are those of the authors and do not represent the views of Wells Fargo.

\bibliography{./main}

\begin{thebibliography}{}

\bibitem[Alber et~al., 2019]{alber2019innvestigation}
Alber, M., Lapuschkin, S., Seegerer, P., H{{\"a}}gele, M., Sch{{\"u}}tt, K.~T.,
  Montavon, G., Samek, W., M{{\"u}}ller, K.-R., D{{\"a}}hne, S., and
  Kindermans, P.-J. (2019).
\newblock innvestigate neural networks!
\newblock {\em Journal of Machine Learning Research}, 20(93):1--8.

\bibitem[Altmann et~al., 2010]{Altmann2010vip}
Altmann, A., Toloşi, L., Sander, O., and Lengauer, T. (2010).
\newblock Permutation importance: a corrected feature importance measure.
\newblock {\em Bioinformatics}, pages Volume 26, Issue 10.

\bibitem[Amplayo et~al., 2019]{Amplayo2019Length}
Amplayo, R.~K., Lim, S., and Hwang, S.-w. (2019).
\newblock Text length adaptation in sentiment classification.
\newblock {\em Proceedings of Machine Learning Research}, pages 101: 1--16.

\bibitem[Arras et~al., 2017]{arras2017relevant}
Arras, L., Horn, F., Montavon, G., M{\"u}ller, K.-R., and Samek, W. (2017).
\newblock ``what is relevant in a text document?'': An interpretable machine
  learning approach.
\newblock {\em PloS one}, 12(8):e0181142.

\bibitem[Bach et~al., 2015]{bach2015pixel}
Bach, S., Binder, A., Montavon, G., Klauschen, F., M{\"u}ller, K.-R., and
  Samek, W. (2015).
\newblock On pixel-wise explanations for non-linear classifier decisions by
  layer-wise relevance propagation.
\newblock {\em PloS one}, 10(7):e0130140.

\bibitem[Bird et~al., 2009]{Bird2009}
Bird, S., Loper, E., and Klein, E. (2009).
\newblock Natural language processing with python.
\newblock {\em O’Reilly Media Inc.}

\bibitem[Blei et~al., 2003]{Blei2003LDA}
Blei, D.~M., Ng, A.~Y., and Jordan, M.~I. (2003).
\newblock Latent dirichlet allocation.
\newblock {\em Journal of Machine Learning Research 3}, pages 993--1022.

\bibitem[Blodgett et~al., 2020]{blodgett2020language}
Blodgett, S.~L., Barocas, S., Daum{\'e}~III, H., and Wallach, H. (2020).
\newblock Language (technology) is power: A critical survey of" bias" in nlp.
\newblock {\em arXiv preprint arXiv:2005.14050}.

\bibitem[Friedman, 2001]{Friedman2001PDP}
Friedman, J. (2001).
\newblock Greedy function approximation: A gradient boosting machine.
\newblock {\em The Annals of Statistics}, pages 29(5):1189--1232.

\bibitem[Garneau et~al., 2019]{Garneau2019OOT}
Garneau, N., Leboeuf, J.-S., and Lamontagne, L. (2019).
\newblock Contextual generation of word embeddings for out of vocabulary words
  in downstream tasks.
\newblock {\em Canadian Conference on AI}.

\bibitem[Gholizadeh and Zhou, 2021]{gholizadeh2021model}
Gholizadeh, S. and Zhou, N. (2021).
\newblock Model explainability in deep learning based natural language
  processing.
\newblock {\em arXiv preprint arXiv:2106.07410}.

\bibitem[Goldstein et~al., 2013]{Goldstein2013ICE}
Goldstein, A., Kapelner, A., Bleich, J., and Pitkin, E. (2013).
\newblock Peeking inside the black box: Visualizing statistical learning with
  plots of individual conditional expectation.
\newblock {\em arXiv preprint arXiv:1309.6392}.

\bibitem[Han and Ghosh, 2020]{Han2020subset}
Han, X. and Ghosh, J. (2020).
\newblock Modelagnostic explanations using minimal forcing subsets.
\newblock {\em arXiv preprint}, page arXiv:2011.00639.

\bibitem[Hu et~al., 2018]{Hu2018LIMEsup}
Hu, L., Chen, J., Nair, V.~N., and Sudjianto, A. (2018).
\newblock Locally interpretable models and effects based on supervised
  partitioning (lime-sup).
\newblock {\em arXiv preprint}, page arXiv:1806.00663.

\bibitem[Khanna et~al., 2019]{Khanna2019black}
Khanna, R., Kim, B., and Ghosh, J. (2019).
\newblock Interpreting black box predictions using fisher kernels.
\newblock In {\em In The 22nd International Conference on Artificial
  Intelligence}, page 3382–3390. PMLR.

\bibitem[Lertvittayakumjorn and Toni, 2021]{Lertvittayakumjorn2021survey}
Lertvittayakumjorn, P. and Toni, F. (2021).
\newblock Explanation-based human debugging of nlp models: A survey.
\newblock {\em arXiv preprint}, page arXiv:2104.15135v3.

\bibitem[Liu, 2012]{Bing2012sent}
Liu, B. (2012).
\newblock Sentiment analysis and opinion mining.
\newblock {\em Morgan \& Claypool Publishers}.

\bibitem[Liu et~al., 2018]{liu2018model}
Liu, X., Chen, J., Nair, V., and Sudjianto, A. (2018).
\newblock Model interpretation: A unified derivative-based framework for
  nonparametric regression and supervised machine learning.
\newblock {\em arXiv preprint arXiv:1808.07216}.

\bibitem[Louppe et~al., 2013]{Louppe2013vip}
Louppe, G., Wehenkel, L., Sutera, A., and Geurts, P. (2013).
\newblock Understanding variable importances in forests of randomized trees.
\newblock In {\em NIPS, Lake Tahoe, United States}.

\bibitem[Madsen et~al., 2021]{madsen2021posthoc}
Madsen, A., Reddy, S., and Chandar, S. (2021).
\newblock Post-hoc interpretability for neural nlp: A survey.
\newblock {\em arXiv preprint arXiv:2108.04840}.

\bibitem[Mohammad, 2020]{mohammad2020practical}
Mohammad, S.~M. (2020).
\newblock Practical and ethical considerations in the effective use of emotion
  and sentiment lexicons.
\newblock {\em arXiv preprint arXiv:2011.03492}.

\bibitem[Nielsen, 2011]{Nielsen2011afinn}
Nielsen, F.~{\AA}. (2011).
\newblock Afinn.
\newblock {\em Informatics and Mathematical Modelling, Technical University of
  Denmark}.

\bibitem[Ribeiro et~al., 2016]{ribeiro2016model}
Ribeiro, M.~T., Singh, S., and Guestrin, C. (2016).
\newblock Model-agnostic interpretability of machine learning.
\newblock {\em arXiv preprint arXiv:1606.05386}.

\bibitem[Silge and Robinson, 2016]{Silge2016tidy}
Silge, J. and Robinson, D. (2016).
\newblock tidytext: Text mining and analysis using tidy data principles in r.
\newblock {\em JOSS}.

\bibitem[Smith-Renner et~al., 2020]{Smith2020exp}
Smith-Renner, A., Fan, R., Birchfield, M., Wu, T., Boyd-Graber, J., Weld,
  D.~S., and Findlater, L. (2020).
\newblock No explainability without accountability: An empirical study of
  explanations and feedback in interactive ml.
\newblock In {\em In Proceedings of the 2020 CHI Conference on Human Factors in
  Computing Systems}, page 1–13.

\end{thebibliography}
\bibliographystyle{apalike}

\end{document}